\documentclass{article}

\usepackage{microtype}
\usepackage{graphicx}
\usepackage{subfigure}
\usepackage{booktabs} 

\usepackage{hyperref}

\usepackage[accepted]{icml2021}

\icmltitlerunning{Learning Fair Policies in Decentralized Cooperative Multi-Agent Reinforcement Learning}

\usepackage{bm}
\usepackage{amssymb}
\usepackage{amsmath}
\usepackage{amsthm}
\usepackage{mathtools}
\newtheorem{theorem}{Theorem}[section]
\newtheorem{lemma}[theorem]{Lemma}
\newtheorem{corollary}[theorem]{Corollary}

\usepackage{musicography}
{\theoremstyle{plain}
\newtheorem{assumption}{Assumption}
\newtheorem*{assumptionbis}{Assumption}
 }

\newcommand*{\prob}{\mathsf{P}}
\newcommand{\St}{{\mathcal S}}
\newcommand{\Ac}{{\mathcal A}}
\newcommand{\Ob}{{\mathcal O}}
\newcommand{\T}{{P}}
\newcommand{\R}{{r}}
\newcommand{\vR}{{\bm r}}
\newcommand{\vvf}{{\bm V}} 
\newcommand{\vQ}{{\bm Q}} %
\newcommand{\vJ}{\bm J} 
\newcommand{\w}{{\bm w}} 
\newcommand{\nO}{D} 
\newcommand{\nA}{N} 
\newcommand{\swf}{\phi}
\DeclareMathOperator{\Var}{Var}

\newcommand{\self}{self-oriented}
\newcommand{\team}{team-oriented}

\newcommand{\err}{\bm{\epsilon}}
\newcommand{\gr}{\mathfrak{g}}
\newcommand{\norm}[1]{\left\lVert#1\right\rVert}

\usepackage{amsmath}
\DeclareMathOperator*{\argmax}{argmax}

\newcommand{\Expect}{\mathbb E}

\usepackage{etoolbox}
\newtoggle{final}
\toggletrue{final}

\newcommand{\paul}[1]{\iftoggle{final}{#1}{{\color{orange} #1}}}

\newcommand{\pw}[1]{{#1}}
\newcommand{\mz}[1]{{#1}}
\newcommand{\us}[1]{{#1}}
\newcommand{\cg}[1]{{#1}}

\usepackage[small]{caption}

\begin{document}

\twocolumn[
\icmltitle{Learning Fair Policies in Decentralized Cooperative \\ Multi-Agent Reinforcement Learning}



\icmlsetsymbol{equal}{*}

\begin{icmlauthorlist}
\icmlauthor{Matthieu Zimmer}{equal,one}
\icmlauthor{Claire Glanois}{equal,one}
\icmlauthor{Umer Siddique}{one}
\icmlauthor{Paul Weng}{one,two}
\end{icmlauthorlist}

\icmlaffiliation{one}{UM-SJTU Joint Institute, Shanghai Jiao Tong University, China}
\icmlaffiliation{two}{Department of Automation, Shanghai Jiao Tong University, Shanghai, China}

\icmlcorrespondingauthor{Paul Weng}{paul.weng@sjtu.edu.cn}

\icmlkeywords{ICML, Fairness, Multi-agent, Decentralized, Actor-Critic}

\vskip 0.3in
]



\printAffiliationsAndNotice{\icmlEqualContribution} 

\begin{abstract}
  We consider the problem of learning fair policies in (deep) cooperative multi-agent reinforcement learning (MARL). 
  We formalize it in a principled way as the problem of optimizing a welfare function that explicitly encodes two important aspects of fairness: efficiency and equity. 
  We provide a theoretical analysis of the convergence of policy gradient for this problem.
  As a solution method, we propose a novel neural network architecture, which is composed of two sub-networks specifically designed for taking into account \pw{these} two aspects of fairness. 
  In experiments, we demonstrate the importance of the two sub-networks for fair optimization. 
  Our overall approach is general as it can accommodate any (sub)differentiable welfare function.
  Therefore, it is compatible with various notions of fairness that have been proposed in the literature (e.g., lexicographic maximin, generalized Gini social welfare function, proportional fairness).
  Our method is generic and can be implemented in various MARL settings: centralized training and decentralized execution, or fully decentralized. 
  Finally, we experimentally validate our approach in various domains and show that it can perform much better than previous methods, \cg{both in terms of efficiency and equity}.
\end{abstract}

\section{Introduction}\label{sec:intro}

Adaptive distributed control systems start to be considered in real applications, e.g., 
traffic light control \citep{Pol2016}, 
multi-robot patrolling \citep{portugalDistributedMultirobotPatrol2013}, or
internet congestion \citep{jayInternetCongestionControl2019}.
Although those systems generally may impact many end-users, the current main focus 
is on their performance with respect to the total (or average) of some per-user efficiency measure (e.g., waiting times of cars in traffic light control, 
safety of different sites in patrolling, or
throughput of users in internet congestion).
However, this approach is clearly \cg{unsatisfactory} due to the users' conflicting interests.
Thus, for such systems, fairness becomes a key factor to consider in their designs for their successful deployments and operations.

Fairness is a multifaceted concept \pw{(Section~\ref{sec:relatedwork})}, which can refer to or include different aspects, e.g., impartiality, 
equity, 
Pareto-efficiency, 
envy-freeness, or
proportionality among others.
Given the importance of this notion, it has been investigated in various scientific disciplines, from philosophy to computer science, including economics and applied mathematics.
In this work, fairness specifically refers to the combination of the first three aspects. 
Interestingly, this definition of fairness can be encoded in a \emph{fair} social welfare function, which combines the users' utilities and can be used to evaluate and compare different solutions.

In this paper, we consider adaptive distributed control systems modeled as cooperative decentralized multi-agent reinforcement learning (MARL), and study the problem of learning fair distributed policies.
This approach applies to situations where a system designer needs to implement a distributed system to solve a specific task (e.g., traffic regulation, patrolling, or congestion control) for many users in a fair way.
Thanks to our definition of fairness, this problem can be expressed as a fair optimization problem, i.e., optimization of a fair social welfare function.

This formalization can then be tackled with standard multi-agent deep reinforcement learning techniques.
Yet, as agents need to learn both efficiency and equity, two conflicting aspects of fairness, a naive approach is insufficient, as shown in our experiments.
Thus, we propose a novel  architecture specifically designed for fair optimization in multi-agent deep reinforcement learning  (MADRL), which is shown to experimentally over-perform previous approaches.

\paragraph{Contributions}
We formulate a general and principled model for the problem of learning fair solutions in cooperative multi-agent reinforcement learning \pw{(Section~\ref{sec:formalization})}.
We propose a simple, scalable and efficient decentralized method to solve this problem \pw{(Section~\ref{sec:method})}.
We also provide a theoretical analysis of the convergence of policy gradient for this problem (Section~\ref{sec:analysis}).
To validate our approach, we extensively compare it with previous approaches and evaluate it on a diverse set of domains \pw{(Section~\ref{sec:experiments})}.

\section{Related Work}\label{sec:relatedwork} 

The notion of fairness has been extensively studied in political philosophy (e.g., \citep{Rawls71}), political sciences (e.g., \citep{bramsFairDivisionCakeCutting1996}) and in economics (e.g., \cite{Moulin04}). 
This literature has led to the considerations of various aspects of fairness, e.g., equal treatment of equals, 
efficiency with respect to Pareto dominance,
equal distribution (of goods, wealth, opportunities...), or envy-freeness, 
which have been exploited in more \pw{applied} fields, such as operations research (OR), artificial intelligence (AI), or machine learning (ML).
In this paper, we follow the approaches based on social welfare functions \citep{
Moulin04}. 
\pw{Various formulations have been considered, e.g., 
the utilitarian one that considers the users' total utility or the egalitarian approach that focuses on the lowest utility.
In this paper, we \cg{investigate} a family of fair social welfare functions that encodes impartiality, equity, and efficiency (see Section~\ref{sec:fairness}).}


Such approaches, \cg{referred to as } \emph{fair optimization}, have been adopted before in OR and related fields \citep{OgryczakLussPioroNaceTomaszewski14},
 \cg{and have} many applications notably in networking \citep{amaldiSinglepathNetworkRouting2013,shiFairnessWirelessNetworks2014}.
Various classic OR problems have been studied in the fair optimization setting, e.g., 
location  \citep{neidhardtDataFusionOptimal2008},
allocation \citep{bertsimasPriceFairness2011},
or Markov decision process \citep{OgryczakPernyWeng13}.
As typical in OR, those works \pw{usually} deal with a \pw{centralized} and known environment setting.
Our work can be seen as an extension of this literature to the decentralized and learning setting.

In AI, fairness has been considered in multi-agent systems with a large focus on resource allocation problems, notably with envy-freeness \citep{chevaleyreIssuesMultiagentResource2006} and some works in non-cooperative games \citep{de-JongTuylsVerbeeck08,HaoLeung16}.
In contrast, we deal with more complex control problems, but in the \textbf{cooperative} setting.
\pw{As such, our proposition is based on MARL instead of a game-theoretic formulation, which is more suitable for the non-cooperative setting}.
Besides, we \cg{formulate} fairness with respect to users instead of agents, which is a more general framework.

Recently, fairness has started to become an important topic in ML.
\pw{Indeed,} as ML models are deployed in various applications (e.g., banking or law enforcement), the decisions made on their outputs may severely impact some \pw{users} due to the presence of bias in data.
Different ML tasks have been \cg{inspected in this regard}, e.g., classification \citep{dworkFairnessAwareness2012,zafarParityPreferencebasedNotions2017,sharifi-malvajerdiAverageIndividualFairness2019}, ranking \citep{singhPolicyLearningFairness2019}, sequential-decision making \citep{Busa-FeketeSzorenyiWengMannor17} or clustering \citep{chierichettiFairClusteringFairlets2017}. 
Most of such work focuses on the impartiality aspect of fairness, expressed at the individual or group level, which leads to a constrained-based or penalty-based formulation.
However, some recent work
\citep{SpeicherHeidariGrgicHlacaGummadiSinglaWellerZafar18,HeidariFerrariGummadiKrause18} advocates a \cg{more complete} approach based on fair social welfare function that we also adopt in 
\pw{our work. 
Besides, such approach was recently investigated in single-agent deep reinforcement learning (RL) \citep{SiddiqueWengZimmer20}.}

Due to the recent successes of deep RL, research on MADRL has become very active
\citep{Hernandez-Leal2019}. 
Different settings have been considered depending on whether training or execution is centralized or not, state observability is partial or not, and communication is allowed or not.
\pw{Some recent} work focuses on tackling problems related to decentralized training \citep{zhangFullyDecentralizedMultiAgent2018}, communication \citep{foersterLearningCommunicateDeep2016,sukhbaatarLearningMultiagentCommunication2016}, coordination  \citep{Pol2016}, or agent modeling \citep{raileanuModelingOthersUsing2018}.
In cooperative MADRL, the usual approach is based on a utilitarian formulation or a unique common reward signal.

However, fairness has been explicitly considered in multi-agent sequential decision-making in some few exceptions \citep{ZhangShah14,JiangLu19}.
\pw{\citet{ZhangShah14} consider} a regularized maxmin egalitarian approach in order to find an equitable solution. 
\pw{Yet}, this may be \cg{deficient} as the solution without the worse-off agent may not be fair.
Also, this work does not consider learning.
In order to learn fair solutions, \pw{\citet{JiangLu19} propose} FEN, a decentralized method using two main ingredients.
First, a gossip algorithm is used to estimate the average utility obtained by all agents. 
Second, the policy of each agent has a hierarchical architecture, where the high level decides to optimize its own utility or not, and the low level is composed of several sub-policies: the first one optimizes the individual reward gathered by the agent, while the others optimize their probability of being selected by the higher level.
\pw{That work has several limitations.
It implicitly assumes that agents have equal access to resources, which may not be true in practice (see Section~\ref{sec:experiments}).
Besides, fairness is implicitly defined with the coefficient of variation (CV)\footnote{The ratio of the standard deviation to the mean.}, an inequality measure \cg{---measuring the dispersion of the utility---} which does not guarantee efficiency.
}

\section{Formalization}\label{sec:formalization}

\paragraph{Notations}
For any natural integer $n$, $[n]$ denotes the set $\{1, \ldots, n\}$.
Vectors, which are column vectors, and matrices are denoted in bold and their components in normal typeface with indices, e.g., $\bm x = (x_1, \ldots, x_n)$.
For \pw{any} set $X$, $\Delta(X)$ denotes the set of probability measures over $X$.

\subsection{Multi-Agent Reinforcement Learning} \label{sec:MADRL} 

Recall that a \pw{decentralized partially observable} Markov decision process can be defined with the following n-\us{t}uple $\big(\St, \Ac = (\Ac_i)_{i \in [\nA]}, (\Ob_i)_{i \in [\nA]}, \T, (\Omega_i)_{i \in [\nA]}, \R, \gamma\big)$ where $\nA$ is the number of agents,
$\St$ is the global state space,
$\Ac_i$ is the action space of agent $i$, 
$\Ob_i$ is the observation space of agent $i$,
$\T : \St \times \Ac \to \Delta(\St)$ is the joint transition function, 
$\Omega_i : \St \to \Delta(\Ob_i)$ is the observation function of agent $i$,  
$\R$ is a joint reward function, and
$\gamma \in (0, 1)$ is a discount factor.

Since the operations of an agent may impact many different users, we extend the previous formulation by redefining the reward function to be vectorial: $\vR : \St \times \Ac \to \mathbb R^\nO$ where $\nO$ is the number of users.
A user can represent an individual or a group \pw{of individuals}.
We denote $\vR = (\R_k)_{k \in [\nO]}$.
As the system is distributed, the agents may not observe the whole reward vector.
At a given time step, an agent  $i$ observes $\vR_{I_i} = (\R_k)_{k \in I_i}$ where $I_i \subseteq [\nO]$.
\mz{Note that the partial observability of rewards does not imply reward independence between agents.
The rewards depend on the state of the whole system and the actions of all agents.}
For ease of presentation of our solution method, we will assume that \cg{the} set $I_i$ is fixed for each agent $i$ and the sets $I_i$'s of all agents form a partition of $[\nO]$.
Our approach can readily be extended to the more general case where the sets $I_i$ and $I_j$ of two agents may have a non-empty intersection.
Note that our formulation is strictly more general than the usual approach where fairness is defined over agents.
By setting $D=N$ and $I_i = \{i\}$, we can recover the usual formulation.

A joint policy can be written as follows $\bm\pi(\bm a | \bm o) = (\pi_{\pw{1}}(a_1 | o_1), \ldots, \pi_{\pw{N}}(a_N | o_N))$.
The individual policy of the $i^\text{th}$ agent is denoted $\pi_i : \Ob_i \to \Delta(\mathcal{A}_i)$ \pw{since} an agent only perceives \pw{its} local observation.
Likewise, since each agent cannot access the whole reward vector nor the joint state, each agent learns an individual state value function $\hat{\bm V}_{I_i} : \Ob_i \to \mathbb{R}^{|I_i|}$ in order to approximate $\vvf_{I_i}(\bm s) = \Expect_{\bm \pi}\left[\sum_{t=1}^{\infty} \gamma^{t-1} \vR_{I_i,t} \mid \pw{\bm s_0 = \bm s}\right]$\pw{, which represents the utilities of users in $I_i$ in state $\bm s$.}

\subsection{Fairness Formulation} 
\label{sec:fairness} 

The notion of fairness we focus on in this paper encompasses three important aspects \citep{adlerWellBeingFairDistribution2012}: impartiality, equity, and  efficiency. 
Impartiality corresponds to the "equal treatment of equals" principle, which is arguably one of the most important pillars of fairness.
In this paper, we assume that all users are identical and should therefore be treated similarly.
In terms of utility vectors, impartiality \pw{implies} that  permutations of a utility vector \pw{are equivalent} solutions.

Equity is based on the \emph{Pigou-Dalton principle} \citep{Pigou12,Dalton20}, which states that a reward transfer from a better-off user to a worse-off
user yields a fairer solution. 
Formally, it is expressed as follows:
for any utility vector $\bm u \in \mathbb R^D$, if $u_j - u_i > \varepsilon > 0$, then $\bm u + \varepsilon \bm e_i - \varepsilon \bm e_j$ is considered \pw{fairer} than $\bm u$, 
where $\bm e_i \in \mathbb R^D$ (resp. $\bm e_j \in \mathbb R^D$) is the null vector except in component $i$ (resp. $j$) where it is equal to $1$.
Such a transfer is called a \emph{Pigou-Dalton transfer}.
This principle formally expresses the notion of equal distribution of "wealth", which is the basis of the equity property that we want our fairness concept to satisfy.
This principle is natural in our context where accumulated rewards vectors are interpreted as wealth distributions.

Efficiency states that between two feasible solutions, if one solution is (weakly or strictly) preferred by all users, then \pw{it} should be preferred to the other one.
This simply corresponds to Pareto dominance\footnote{For any $(\bm u, \bm u') \in \mathbb R^{\nO\times\nO}$, $\bm u$ Pareto-dominates $\bm u'$ (denoted $\bm u \succ \bm u'$) if $\forall i, u_i \ge u_i'$ and $\exists j, u_j > u'_j$.} in the space of users' utilities.
Although efficiency is not always considered an integral part of fairness, one could argue that it would be unfair in the name of equity not to increase the rewards of all or some users while not decreasing the rewards of any other users, if that were possible.
Without efficiency, giving no reward to all users would be as good as giving 100 to all users.


To make this notion of fairness operational, we adopt the approach based on social welfare functions.
A \emph{social welfare function} (SWF) is a function $\swf : \mathbb R^D \to \mathbb R$, which aggregates a utility vector and measures how good it is in terms of social good.
Naturally, among all SWFs, we consider those that satisfy the notion of fairness we have just discussed.

Impartiality implies that an SWF $\phi$ should be symmetric, that is $\phi$ should be independent of the order of its arguments, i.e., $\phi(\bm u) = \phi(\bm u_\sigma)$ where $\sigma$ is a permutation and $\bm u_\sigma$ is the vector obtained from vector $\bm u$ permuted by $\sigma$. 
Efficiency means that $\phi$ should be strictly monotonic with respect to Pareto dominance, i.e., $\bm u \succ \bm u' \Rightarrow \phi(\bm u) > \phi(\bm u')$. 
Finally, the Pigou-Dalton principle implies that $\phi$ should be strictly Schur-concave (i.e., strictly monotonic with respect to Pigou-Dalton transfers).  

In this paper, an SWF will be called \emph{fair SWF} if it satisfies the three previous properties.
Many fair SWFs have been proposed in the literature.
One may distinguish two main families.
The first is the generalized Gini SWF (GGF), which is defined as follows:
\begin{align}\label{eq:gini}
    G_{\bm w}(\bm u) = \sum_{k \in [D]} w_k u_k^\uparrow
\end{align}
where $\bm w \in [0, 1]^D$ is a fixed strictly decreasing weight vector (i.e., $w_1 > w_2 > \ldots > w_D$) and 
$\bm u^\uparrow$ is the vector obtained from $\bm u$ by sorting its components in an increasing order.
By choosing appropriately the weights $\bm w$ (and in some cases allowing them to be weakly decreasing), this family of SWF includes the maxmin egalitarian approach ($w_1 = 1, w_2=\ldots=w_D=0$), the regularized maxmin egalitarian approach ($w_1 = 1, w_2=\ldots=w_D=\varepsilon$), the leximin egalitarian approach ($\forall i, w_i/w_{i+1} \to \infty$), or the utilitarian approach ($\forall i, w_i=1$).
However, requiring that weights $\bm w$ are strictly decreasing is important to ensure that the obtained solution is fair (i.e., Pareto optimal and equitable).
Another family of SWF can be written as follows:
\begin{align*}
    \phi(\bm u) = \sum_{k \in [D]} U(u_k) 
\end{align*}
where $U : \mathbb R \to \mathbb R$ is strictly increasing and strictly concave.
Recall a function that is symmetric and strictly concave is strictly Schur-concave.
This family is very general and includes proportional fairness \citep{PioroMalicskoFodor02} when $U(x) = \log(x)$ and more generally 
$\alpha$-fairness \citep{MoWalrand00} when $U_\alpha(x) = \frac{x^{1-\alpha}}{1-\alpha}$ if $\alpha \neq 1$ and $U_\alpha(x) = \log(x)$ otherwise, with parameter $\alpha > 0$ controlling the aversion to inequality.
We denote the corresponding SWF $\phi_\alpha$.
When $\alpha \to \infty$, it tends to the leximin egalitarian formulation.
Even more broadly, this family includes SWFs derived from the generalized entropy index \citep{Shorrocks80}.

The exact choice of an SWF depends on the specific problem one wants to solve.
Intuitively, \pw{a fair SWF trades off between equity and efficiency:  optimizing a given fair SWF amounts to selecting} among Pareto-optimal solutions the one with the best trade-off.
As we aim at designing a generic approach for fairness, we leave this choice unspecified.

\subsection{Problem Statement} 
\label{sec:problem} 

As usual, in order to tackle problems with large-sized or even continuous state/action spaces, we assume that the policy space is parameterized.
Based on the notion of fair SWF, our problem can be simply formulated as follows:
\begin{align} \label{eq:opticen}
    \max_{\bm \theta} \swf(\vJ(\bm \theta))
\end{align}
where $\bm \theta$ is the parameters of the joint policy of all the agents and $J_k(\bm \theta) = \mathbb E_{\bm\theta}[\sum_t \gamma^t \R_{k,t}]$ is the expected sum of discounted rewards for user $k$.
Interestingly, the two families of fair SWFs that we recalled correspond to concave functions, which implies that  \eqref{eq:opticen} is a convex optimization problem.

Since each user's utility only depends on one agent, our problem can be written as:
\begin{align} \label{eq:optidis}
    \max_{\bm \theta} \mathfrak{J}(\bm\theta) = \max_{\bm \theta} \swf(\vJ_{I_1}(\theta_1), \ldots, \vJ_{I_N}(\theta_N))
\end{align}
where $\bm \theta = (\theta_1, \ldots, \theta_N)$ is the policies' parameters of $\bm\pi = (\pi_1, \ldots, \pi_N)$ respectively and $I_i$ corresponds to the set of indices of users whose utilities depend on agent $i$.
In the next section, we propose an efficient MADRL method to solve this problem.
Note that although $\vJ(\bm\theta)$ is vectorial, this is a \textbf{single} objective optimization problem since $\phi : \mathbb R^D \to \mathbb R$.
We leave for future work the case where the satisfaction of one user may depend on several agents.

This formulation, which may appear restrictive, is already a generalization of the usual setting where fairness is defined over agents.
\pw{Moreover, i}t enjoys \cg{attractive} advantages.
It is simple, and \cg{transparent}, \cg{openly presenting} what is optimized.
It is theoretically-founded as fair SWFs encode a clear and well-defined notion of fairness.
This formulation and our solution method are generic, since it accepts any (sub)differentiable social welfare function (actually, even if it does not encode fairness).

\section{Solution Method} 
\label{sec:method}  

\mz{To learn distributed fair policies, our solution is based on the optimization of SWFs combined with communication between agents. 
To efficiently optimize the SWF, we propose Self-Oriented Team-Oriented networks (SOTO) updated by dedicated policy gradients (Algorithm \ref{algo:soto}).
}

\subsection{Policy Gradient} \label{sec:pg}
As the problem we want to solve can be expressed as a convex optimization problem, we adopt a policy gradient approach implemented in an actor-critic architecture for \cg{increased} efficiency.
\cg{In the context of} \pw{decentralized} policies, we can derive a direction to optimize the SWF for the $i^\text{th}$ agent (see \eqref{eq:optidis}):
\begin{equation}
\nabla_{\theta_i} \swf (\vJ({\bm\theta})) = {\nabla_{\bm u} \swf (\vJ({\bm\theta}))}^\intercal \cdot \nabla_{\theta_i} \vJ({\bm\theta}), \label{eq:swfpolgrad}
\end{equation}
where $\nabla_{\theta_i} \vJ({\bm\theta}) $ is a $D \times |\theta_i|$-matrix representing the usual stochastic policy gradient over the $D$ different reward components and $\nabla_{\bm u} \swf (\vJ({\bm\theta}))$ is a $D$-dimensional vector. 
For instance, with GGF, $\nabla_{\bm u} G_{\bm w}(\vJ({\bm\theta})) = \w_\sigma$ where $\sigma$ is a permutation that sorts $\vJ({\bm\theta})$ in an increasing order.
Similarly, for $\alpha$-fairness, we have $\nabla_{\bm u} \phi_\alpha (\vJ({\bm\theta})) = \vJ({\bm\theta})^{-\alpha}$ where exponentiation is componentwise.

\pw{Let} ${\bm A}(\bm s, \bm a)$ \pw{denote} a $D$-dimensional vector representing the joint advantage function of taking the joint action $\bm a$ in joint state $\bm s$ under the joint policy parameterized by ${\bm\theta}$.
Using the policy gradient theorem \citep{SuttonMcAllesterSinghMansour00} and since the policies are independent, the gradient can be written as: 
\begin{align}
    \nabla_{\theta_i} \vJ({\bm\theta})
    =& \ \mathbb{E}_{\bm \theta} \Big[ {\bm A}(\bm s, \bm a) \cdot \nabla_{\theta_i} \log {\bm \pi_{\bm\theta}(\bm a | \bm s)}^\intercal \Big] \label{eq:swfpolgrad2} \\
    \approx& \ \mathbb{E}_{\bm \theta} \Big[ \Big( {\bm A}_{I_j}(\bm s, \bm a) \Big)_{j \in [N] } \cdot { \nabla_{\theta_i} \log \pi_{\theta_i}(a_i|o_i) }^\intercal \Big], \notag
\end{align}
where $a_i$ refers to the individual action taken by the $i^{\text{th}}$ agent, $o_i$ is the local observation of the $i^{\text{th}}$ agent sampled from $\Omega_i(\bm s)$ and $\nabla_{\theta_i} \log \pi_{\theta_i}(a_i|o_i)$ is a $|\theta_i|$-dimensional vector.
The approximation is due to using decentralized policies with local observations.

However, in the decentralized multi-agent setting, computing $( {\bm A}_{I_j}(\bm s, \bm a) )_{j \in [N] }$ would usually require a centralized critic, thus computing the correct direction $\nabla_{\theta_i} \swf (\vJ({\bm\theta}))$ is generally not possible.
Instead, to approximate the aggregated advantages, we use the local critic of each agent (each critic ignores the effects of other agents):
\begin{equation}\label{eq:tderror}
    {\bm A}_{I_i}(\bm s,\bm a) \approx \hat{\bm A}_{I_i}(o_i, a_i) = \vR_{I_i} + \gamma \hat{\bm V}_{I_i}(o'_i) - \hat{\bm V}_{I_i}(o_i),
\end{equation}
with $\bm s' \sim P(\cdot | \bm s, \bm a) $ and $o'_i \sim \Omega_i(\bm s')$.
Hence, to approximate the aggregate advantage $( {\bm A}_{I_j}(\bm s, \bm a) )_{j \in [N] }$, the agents share their local advantages $\hat{\bm A}_{I_j}(o_j, a_j)$.
We denote this approximated aggregate advantage by $\hat{\bm A}(\bm o, \bm a)$: 
\begin{align} \label{eq:advapprox}
    \Big( {\bm A}_{I_j}(\bm s, \bm a) \Big)_{j \in [N] }\!\!
    \approx \!
    \hat{\bm A}(\bm o, \bm a)
    =
    \Big(   \hat{\bm A}_{I_j}(o_j, a_j) \Big)_{j \in [N] }
\end{align}
In practice, instead of using the temporal difference \eqref{eq:tderror} over one transition, TD($\lambda$) can be used to reduce the bias of this estimation \citep{Sutton2018,SchulmanMoritzLevineJordanAbbeel16}.



By combining \eqref{eq:swfpolgrad}, \eqref{eq:swfpolgrad2} and \eqref{eq:advapprox}, the SWF policy gradient direction becomes: 
\begin{equation} \label{eq:swfgradfinal}
      \nabla_{\theta_i} \swf (\vJ({\bm\theta})) \approx \mathbb{E}_{\bm \theta} \Big[ \hat{\bm A}^\text{SWF} \cdot { \nabla_{\theta_i} \log \pi_{\theta_i}(a_i|o_i) }^\intercal \Big], 
\end{equation}
where $\hat{\bm A}^\text{SWF} = {\nabla_{\bm u} \swf (\hat\vJ({\bm\theta}))}^\intercal \cdot \hat{\bm A}(\bm o, \bm a)$. 
As the policies are represented by neural networks, this gradient \eqref{eq:swfgradfinal} is convenient to compute
by simply backpropagating $ \hat{\bm A}^\text{SWF}$ inside the policy network.

\begin{figure}[t]
    \centering
    \includegraphics[width=\linewidth]{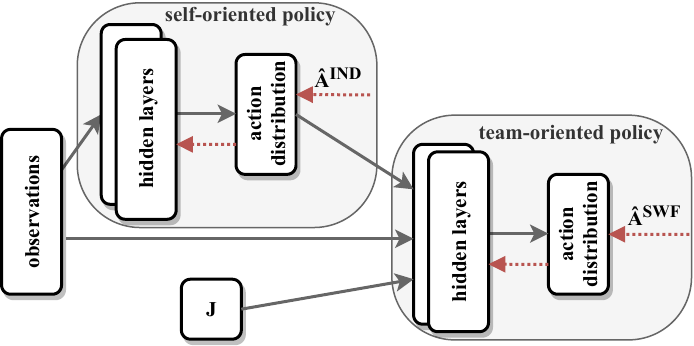}
    \caption{\mz{The SOTO architecture} 
     is composed of a \self{} policy and a \team{} policy.
    The \self{} policy optimizes its individual utility $\vJ_{I_i}$ and recommends an action distribution to the \team{} policy\pw{, which} optimizes the SWF $\phi(\vJ)$.
    Dashed arrows represent backpropagation flow. 
    }
    \label{fig:nnarchitecture}
\end{figure}

\subsection{Neural Network Architecture} \label{sec:NNarchitecture}

Since the agents do not have access to a centralized critic, they may receive conflicting information about the quality of their behaviors from $\hat{\bm A}^\text{SWF}$.
This can prevent an agent $i$ from knowing whether any good/bad performance with respect to its ``individual utility'' $\vJ_{I_i}$ (\self{} performance) or with respect to the global social welfare (\team{} performance) comes from themselves or from the behavior of others (credit assignment problem with non-stationarity).

To avoid this conflict and potential catastrophic forgetting of a good \self{} behavior, we propose a neural network architecture where the policy optimizing the individual utility is no longer disturbed by the local critics of other agents (Figure \ref{fig:nnarchitecture}).

In this architecture, the actor is composed of two sub-networks, which can be viewed as two different policies: one is \self{} and the other \team.
The critic is designed in a similar fashion with two corresponding sub-networks, which take the same inputs as their respective policies, providing a critic to them.
The \self{} policy optimizes its individual utility given by its own critic without taking into account the shared advantages.
The backpropagated advantages $\hat{\bm A}^\text{IND}_i$ for the \self{} policy are defined as: 
\begin{align} \label{eq:selforientedadv}
    \hat{\bm A}^\text{IND}_i = {\nabla_{\bm u_{\bm I_i}} \swf (\vJ({\bm\theta}))}^\intercal \cdot  \hat{\bm A}_{I_i}(o_i, a_i).
\end{align}
Note that for the specific case of $|I_i|=1$, $\hat{\bm A}^\text{IND}_i = \hat{\bm A}_{I_i}(o_i, a_i)$ is used directly as $\nabla_{\bm u_{\bm I_i}} \swf (\vJ({\bm\theta}))$ reduces to scaling the learning rate.
The \team{} policy is updated by \eqref{eq:swfgradfinal} with the aggregated advantages $\hat{\bm A}^\text{SWF}$.

The \team{} policy \pw{takes as input} the distribution proposed by the \self{} one, the estimated $\vJ(\bm\theta)$ of its neighbors, and the usual environmental observations.
Observing $\vJ(\bm\theta)$ is important because it is an essential information for making a fair decision.
Without it, an agent cannot know whether all users are treated fairly or not, since it only observes the rewards of a subset of the users.
For instance, in a resource collection task, an agent needs to know its own score and its neighbors' scores to decide if it wants to start sharing resources. 
Moreover, having access to the output of the \self{} policy greatly simplifies decision-making when an agent's score is lower than its neighbors': it can simply mimic the \self{} policy.

\subsection{Training Schedule} \label{sec:traininschedule}
Because the decentralized execution of independent policies causes non-stationarity in the gathered experience, like previous MADRL methods \citep{foersterLearningCommunicateDeep2016,JiangLu19}, we avoid off-policy learning.
Therefore, to ensure on-policyness of transitions used to train our proposed architecture, the policy applied in the environment must be fixed during a period (we used the minibatch size).
\pw{When an episode starts (see Alg.~\ref{algo:soto}), each agent chooses with probability $\beta$ if it applies its self-oriented policy (or its team-oriented policy otherwise).
Since, an agent must already know how to exploit its own utility before being fair, 
$\beta$ should be high at the beginning of training}.
However, since we ultimately want to optimize the SWF, $\beta$ should decrease \pw{over time}.
When $\beta$ reaches zero, which happens at half of the learning with linear annealing in our experiments, the weights of the \self{} policy will not be updated anymore.

\begin{algorithm}[tb]
   \caption{SOTO algorithm in CLDE scenario}
   \label{algo:soto}
\begin{algorithmic}
\STATE Given $E$ the total number of episode\\
\STATE Initialize $\pi_i, \pi_{i}', v_i, v'_i$, respectively the \team{}/\self{} policies, \team{}/\self{} critics.\\
\FOR{each episode e}
    \STATE $\beta = \text{max}(1 - \frac{e}{0.5 E}, 0) $\\
    \FOR{each agent i}
        \STATE Initialize $\vJ_{I_i} = \bm 0$\\
        \STATE $(\mu_i, w_i) \leftarrow \begin{cases} 
            (\pi_i', v'_i)\text{ with probability } \beta \\
            (\pi_{i}, v_i)\text{ otherwise}
            \end{cases}$  \\
    \ENDFOR
    
    \WHILE{episode e is not completed}
        \STATE Collect $M$ a minibatch of transitions with $\bm \mu$ while updating and sharing $\vJ$ to the neighbors\\
        
        \FOR{each agent i}
            \STATE Update $w_i$ with TD($\lambda$) on $M$\\
            \STATE Compute $\hat{\bm A}_{I_i}(o_i, a_i)$ on $M$
            with $w_i$ and TD($\lambda$) and send it to everyone \hfill \eqref{eq:tderror}\\
            \IF{$\mu_i = \pi_i'$}
                \STATE Update $\pi_i'$ with $\hat{\bm A}^\text{IND}_i$ \hfill \eqref{eq:selforientedadv}
                \ELSE 
                \STATE Collect and form $\hat{\bm A}(\bm o, \bm a)$ \hfill \eqref{eq:advapprox}\\
                \STATE Update $\pi_i$ with $\hat{\bm A}^\text{SWF}$ \hfill \eqref{eq:swfgradfinal}
            \ENDIF
                
            $(\mu_i, w_i) \leftarrow \begin{cases} 
            (\pi_i', v'_i)\text{ with prob. } \beta \\
            (\pi_{i}, v_i)\text{ otherwise}
            \end{cases}$  \\
        \ENDFOR
    \ENDWHILE
\ENDFOR
\end{algorithmic}
\end{algorithm}

\subsection{Communication} \label{sec:communication}

\pw{
The presentation of our method corresponds to the 
Centralized Learning with Decentralized Execution (CLDE) scenario.
We can also evaluate our approach in the Fully Decentralized (FD) scenario.
Recall that for both scenarios, during the execution phase, the communication for an agent $i$ is restricted to the sharing of its $\vJ_{I_i}(\bm \theta)$ with its neighbors.
During learning, while the agents in CLDE can communicate with all other agents, the agents in FD are  allowed to communicate only with neighbors. 
}

Note that our method \pw{never learns a centralized critic:
it} neither communicates full states nor actions, but only $\vJ_{I_i}({\bm\theta})$ and advantages $\hat{\bm A}_{I_i}$. 
Thus, it scales well since the costliest operation depending on $D$ is a matrix product of size $D \times m$ where $m$ is the minibatch size to compute $ \hat{\bm A}^\text{SWF}$.

When a complete minibatch is collected, the advantages are shared during the learning phase to form $\hat{\bm A}(\bm o, \bm a)$. 
\mz{This is only possible in the CLDE scenario.}
In the FD scenario, several rows of the advantages inside $\hat{\bm A}(\bm o, \bm a) $ might be set to zero for agents not being in the neighborhood.
\mz{Instead of using \eqref{eq:advapprox}, we have 
\begin{equation*}
    \hat{\bm A}(\bm o, \bm a)
    =
    \Bigg( \begin{cases}
\hat{\bm A}_{I_j}(o_j, a_j), & \text{if $j \in \mathcal{N}(i)$} \\
\bm 0 & \text{otherwise}
\end{cases}
     \Bigg)_{j \in [N] }, 
\end{equation*}
where $\mathcal{N}(i)$ refers to the neighbors of the $i^\text{th}$ agent.}

\mz{In the following paragraph, we compare the number of messages (1 float) sent by an agent for different algorithms.}
Given $k_i$ the number of neighbors of agent $i$, with $k_i \le N - 1$, at each time step the agent shares its $\vJ_{I_i}({\bm\theta})$ with each neighbor. 
During the update phase, which happens when a minibatch \mz{of size $M$} is full, they also share their estimated advantages for this minibatch \mz{and their estimated $\vJ_{I_i}$ to all agents}. 
Thus, our method sends on average $(k_i + \mz{(1+\frac{1}{M})}(N-1))|I_i|$ messages per step for each agent in the CLDE scenario
and $2k_i|I_i|$ messages in the FD scenario.
As comparison, an agent in FEN sends on average $g\tilde{k}$ messages where $g$ is the number of gossip rounds and $\tilde{k}$ is the number of random chosen agents to send the message.
\pw{Generally, $g$ needs to be greater than the diameter of the graph (so that the information can traverse the graph), which is upperbounded by the number of agents.
\pw{Thus}, assuming $\tilde{k} \approx k_i$,} our method can be \pw{much} more parsimonious than FEN in terms of communication \pw{when $N$ becomes large}. Likewise, a centralized critic would require even more communication by sending on average $\frac{N-1}{N}(\text{dim}(\Ac_i)$ $+$ $\text{dim}(\Ob_i)$ $+$ $|I_i|$ $+$ $1)$ $+$ $k_i|I_i| $ messages every steps.

\section{Theoretical Analysis} 
\label{sec:analysis} 

We analyze the convergence of a policy gradient method to solve Problem~\ref{eq:optidis} under standard assumptions.
The novelty of our analysis is two-fold.
Contrary to previous work, we consider the partial observability of states and rewards, which is more realistic and fits better the decentralized setting.
Besides, the overall objective is a non-linear concave function of the vector of the expected discounted rewards.

We can prove the following convergence result, which we state informally (see Appendix~\ref{app:convergence} for full details):
\begin{theorem}
Under standard assumptions, the SWF objective $\mathfrak{J}(\bm\theta^{k})$ converges almost surely and with a sublinear convergence rate within  a radius of convergence $\tilde{\mathfrak{r}}$ of the optimal value $\mathfrak{J}^{\ast}$ 
where $\tilde{\mathfrak{r}}$ depend on the approximation errors of (a) estimating $\vJ$, (b) estimating $\bm A(\bm o, \bm a)$, and (c) ignoring the effects of one agent's action over other agents.
\end{theorem}
Interestingly, this result implies a corollary, which provides a high-probability bound on the number of iteration steps before convergence.
We provide all the details, further discussion, and the proofs in Appendix~\ref{app:convergence}.
Besides, this theoretical analysis somewhat further justifies our architecture with a specific critic for the self-oriented policy, which helps reduce error (b) and therefore the radius of convergence.

\section{Experiments} 
\label{sec:experiments} 

\mz{
To test our algorithms, we carried out experiments in three different domains (detailed descriptions is available in the appendix): Matthew Effect \citep{JiangLu19}, distributed traffic light control \citep{SUMO2018} and distributed data center control \citep{iroko}.
We also evaluated our approach on the two other domains proposed by \citet{JiangLu19}, Job Scheduling and Plant Manufacturing. However, Job Scheduling \cg{being an} easy artificial domain and Plant Manufacturing \cg{having} an artificially-designed reward function, \us{most of} those results are 
presented in the appendix.}

\us{The first domain is} 
Matthew Effect where 10 pac-men with different initial sizes have to collect resources which reappear randomly each time they are collected in a grid. The more \pw{resources} an agent collects, the easier the task becomes \mz{for an agent} because \pw{its} size and speed increase.

\us{The second domain} \cg{adopted is a} 
distributed traffic light control \cg{scenario}. \us{In this problem,} we simulate a 3x3 intersection grid with Simulation of Urban Mobility (SUMO) 
where each \mz{of the nine agents} controls the traffic light phase of one intersection.
\us{The global state is composed of \pw{the} waiting time, density of cars, queue lengths\pw{, and current traffic-light phase} of each intersection.
For each agent, an action \cg{amounts} to choos\pw{ing} the next \pw{traffic-light} phase.}
The reward function \pw{of an agent} is defined as the negative total waiting time at \pw{its} intersection.
Fairness can be understood as having low waiting times at every intersection.
Note this domain is typically an example where there is no equal access to resources: some intersections will have naturally more traffic than others. 

\us{
Our third domain is a data center control problem, where 16 hosts are connected with 20 switches in a fat-tree topology (\pw{see Figure~\ref{fig:iroko_network}} in the appendix).
\pw{The network is shared by a certain number of hosts.}
The state is composed of information statistics about network features and the goal of each host/agent is to minimize the queue lengths in network switches.
\mz{The continuous action correspond\pw{s} to the allowed bandwidth for a host.}
}

In all our experiments, we rely on the Proximal Policy Optimization (PPO) algorithm \citep{SchulmanWolskiDhariwalRadfordKlimov17}.
The detailed hyperparameters are provided and in Appendix~\ref{app:hyper} and available online\footnote{\url{https://gitlab.com/AAAL/DFRL}}.
To \pw{demonstrate} the generality of our approach, we run our method with GGF (using $\w_i = \frac{1}{2^{i}}$) and $\alpha$-fairness (using $\alpha = 0.9$).
The different statistics are computed over the 50 last trajectories of 5 different runs.

We name the methods that we evaluate as follows. 
Basic($G_{\bm w}$) and Basic($\phi_\alpha$) refer to baselines where PPO optimizes the SWF directly without our proposed neural network architecture.
It is equivalent \pw{to} removing the \pw{self-oriented} policy from our architecture and keeping the observation of neighbors $\vJ({\bm \theta})$. 
SOTO($G_{\bm w}$) and SOTO($\phi_\alpha$) refer to instances of our proposed method.
The prefix "FD" refers to the fully decentralized version.
\us{We also compare our methods with state-of-art algorithms such as FEN \citep{JiangLu19}, a centralized critic method COMA \citep{foerster2017counterfactual} and value-based algorithm WQMIX \citep{rashid2020weighted}. FEN without gossip\mz{, labeled "FEN-g"}, assumes that the agents know the average utility (e.g., by exchanging all their utilities).}

\paragraph{\pw{How does our architecture SOTO perform?}}

\begin{figure}[t]
    \centering
    \includegraphics[width=\linewidth]{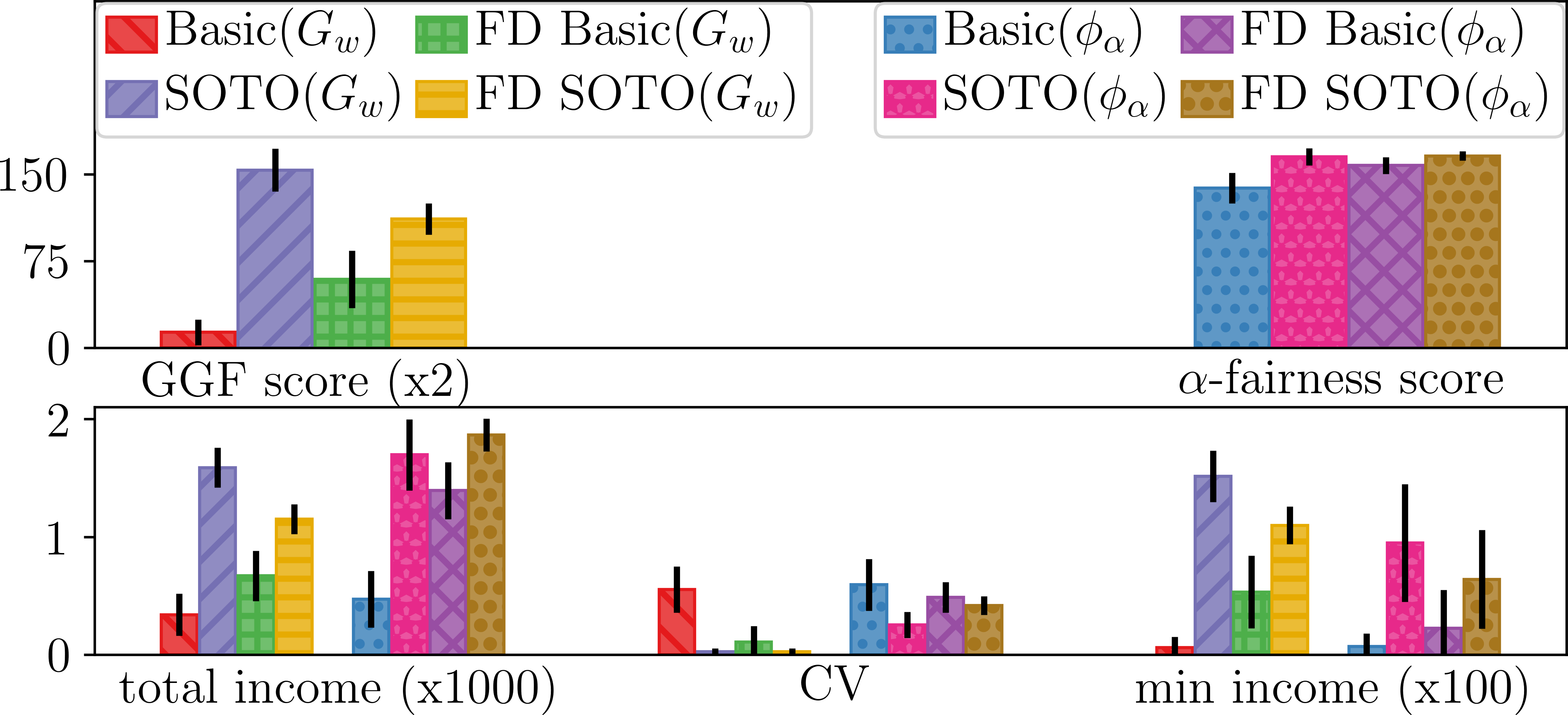}
    \caption{Comparison of SOTO and Basic in Matthew Effect in the CLDE and FD scenarios with GGF and $\alpha$-fairness.}
    \label{fig:arch_vs_basic}
\end{figure}

\pw{We first discuss}
\us{the experimental results} in Matthew Effect and compare our architecture SOTO with several baselines.
\pw{Similar observations can be made in other domains.}
Comparing to Basic, Figure~\ref{fig:arch_vs_basic} shows that
SOTO provides a large improvement over the different criteria (\pw{for CV lower is better}).
Both in CLDE and FD scenarios, with GGF and $\alpha$-fairness, our architecture Pareto-dominates the equivalent approach using the basic architecture without it.

\begin{figure}[t]
    \centering
    \includegraphics[width=\linewidth]{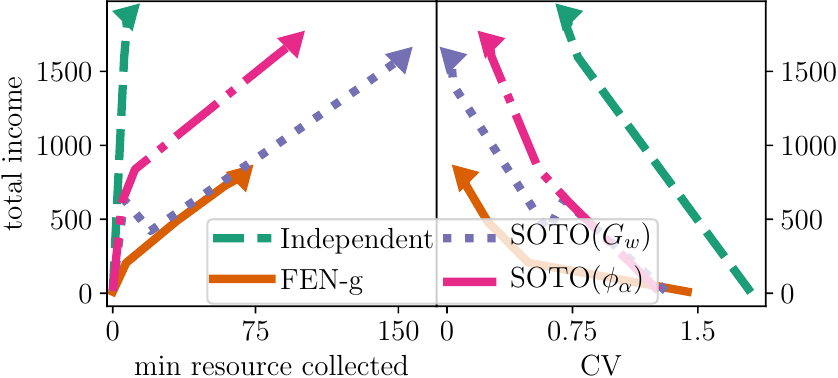}
    \caption{Trajectory of solutions reached by Independent, FEN, and SOTO on Matthew Effect in the CLDE scenario.}
    \label{fig:results_matthew_pareto}
\end{figure}

\begin{figure}[t]
    \centering
    \includegraphics[width=\linewidth]{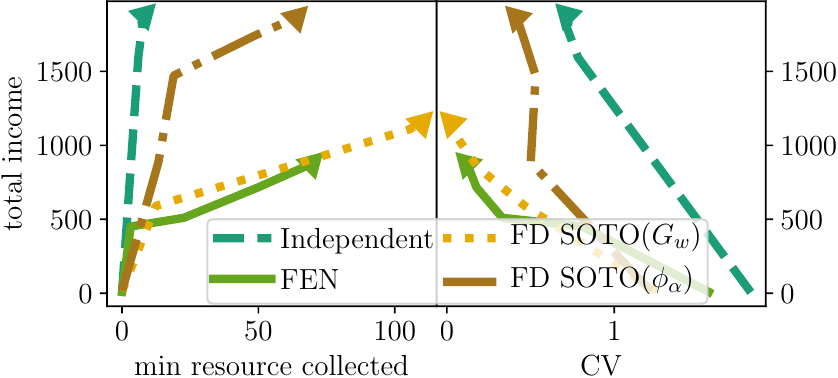}
    \caption{Trajectory of solutions reached by Independent, FEN, and SOTO on Matthew Effect in the FD scenario.}
    \label{fig:results_matthew_pareto_fd}
\end{figure}

To compare with FEN, for better legibility, we plot the trajectories of the policies obtained during training in the space "efficiency" vs "equity" (Figure~\ref{fig:results_matthew_pareto} and Figure~\ref{fig:results_matthew_pareto_fd} in the CLDE and the FD scenario respectively):
total income vs min resource collected (higher in both dimensions is better) and total income vs CV (lower CV is better).
As a sanity check, we include Independent where each agent optimizes its own utility, which in this domain is similar to optimizing the total income.
This plot clearly shows that FEN converges to a worse policy  than SOTO($G_{\bm w}$) both in terms of efficiency and equity. 
SOTO($\phi_\alpha$) can also Pareto-dominate FEN in terms of min resource collected, but not in terms of CV.
These plots illustrate that $\alpha$-fairness provides a different trade-off between efficiency and equity compared to GGF.
Our experiments also suggest that FEN may perform well in terms of CV due to its low efficiency. 

\mz{
As a sanity check, we also compare our method \us{with state-of-the-art standard algorithms such as COMA and WQMIX.}  
In addition, we consider two other variants:
COMA($G_{\bm w}$), which is COMA extended to optimize $G_{\bm w}$; 
CC($G_{\bm w}$), which is the equivalent of Basic \us{but} with a centralized state-value function.
Note that because of negative rewards, applying $\phi_\alpha$ is not possible without tuning the reward function.

\begin{figure}[t]
    \centering
    \includegraphics[width=0.95\linewidth]{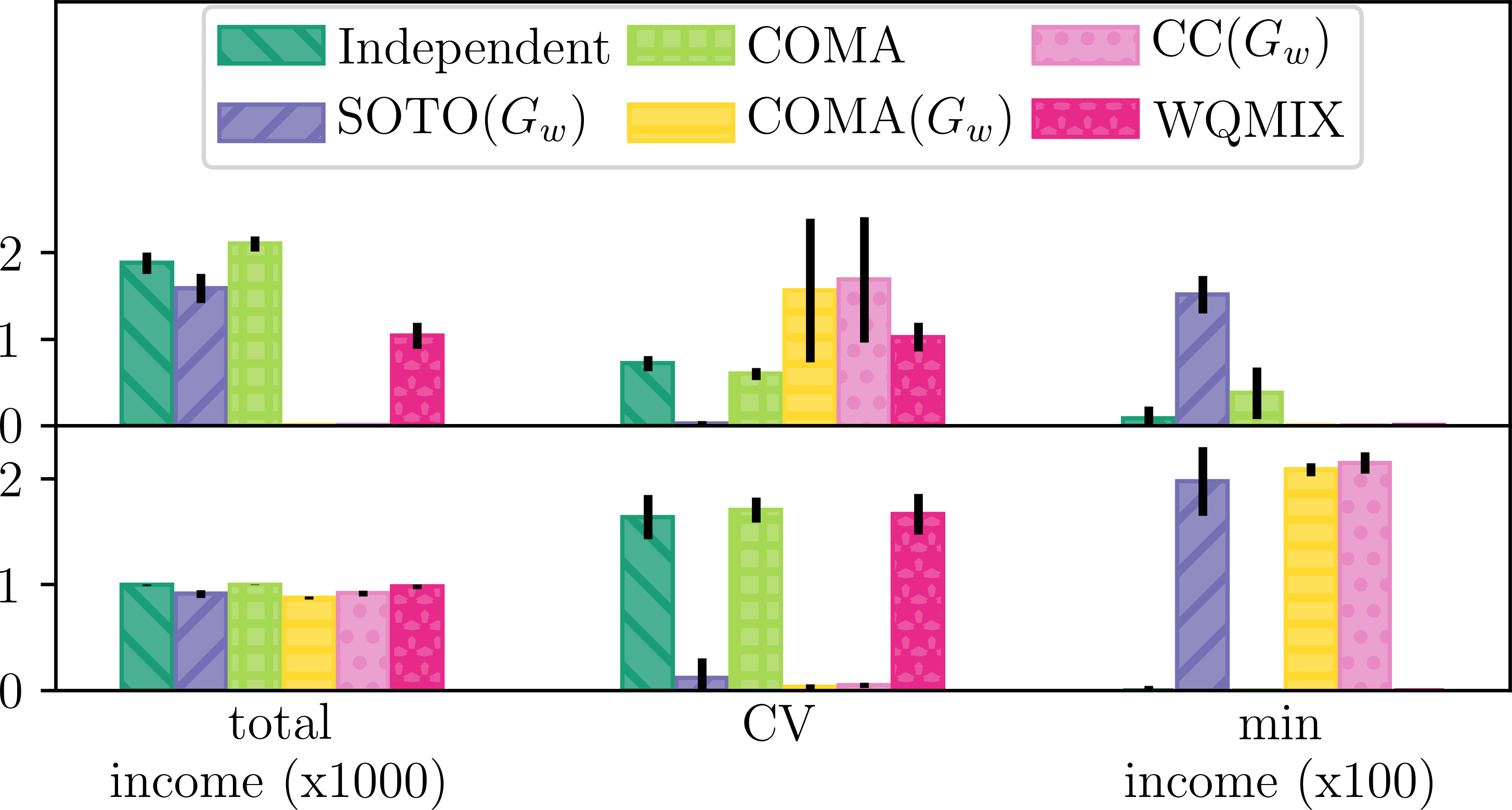}
    \caption{Comparison of SOTO, centralized critic\us{, value based approaches} and Independent in Matthew Effect (top) and Job Scheduling (bottom) in the CLDE scenario. 
    }
    \label{fig:centralizedcritic}
\end{figure}

\begin{figure}[t]
    \centering
    \includegraphics[width=\linewidth]{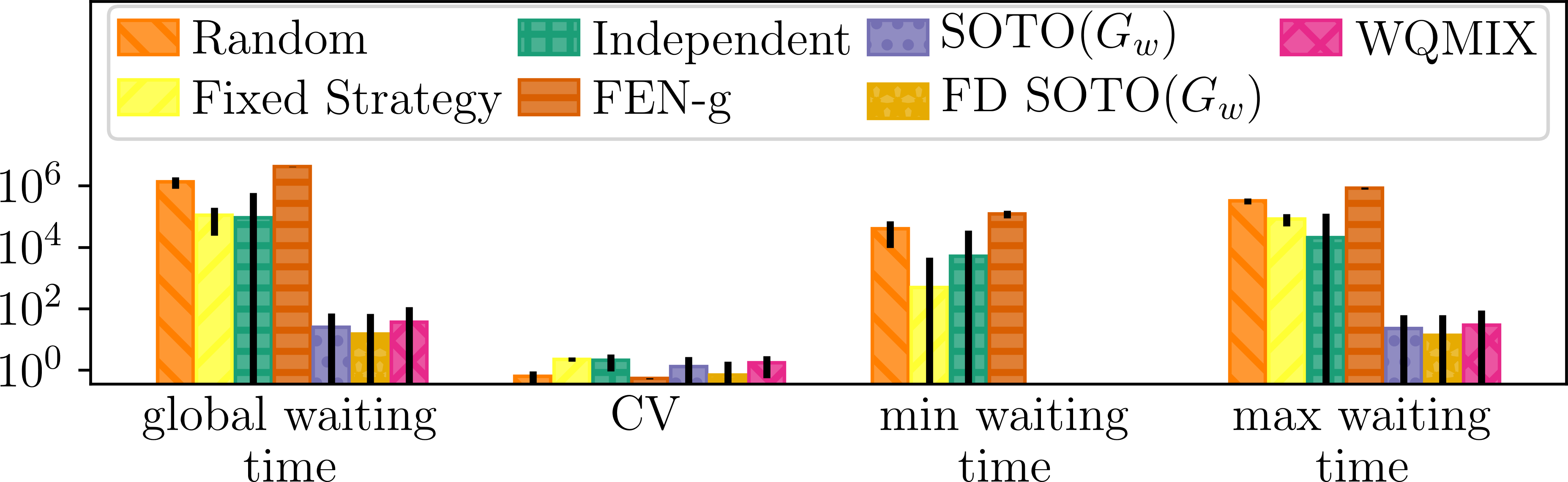}
    \caption{Algorithms' performances in the SUMO environment.}
    \label{fig:sumo_perf}
\end{figure}

\begin{figure}[t]
    \centering
    \includegraphics[width=0.9\linewidth]{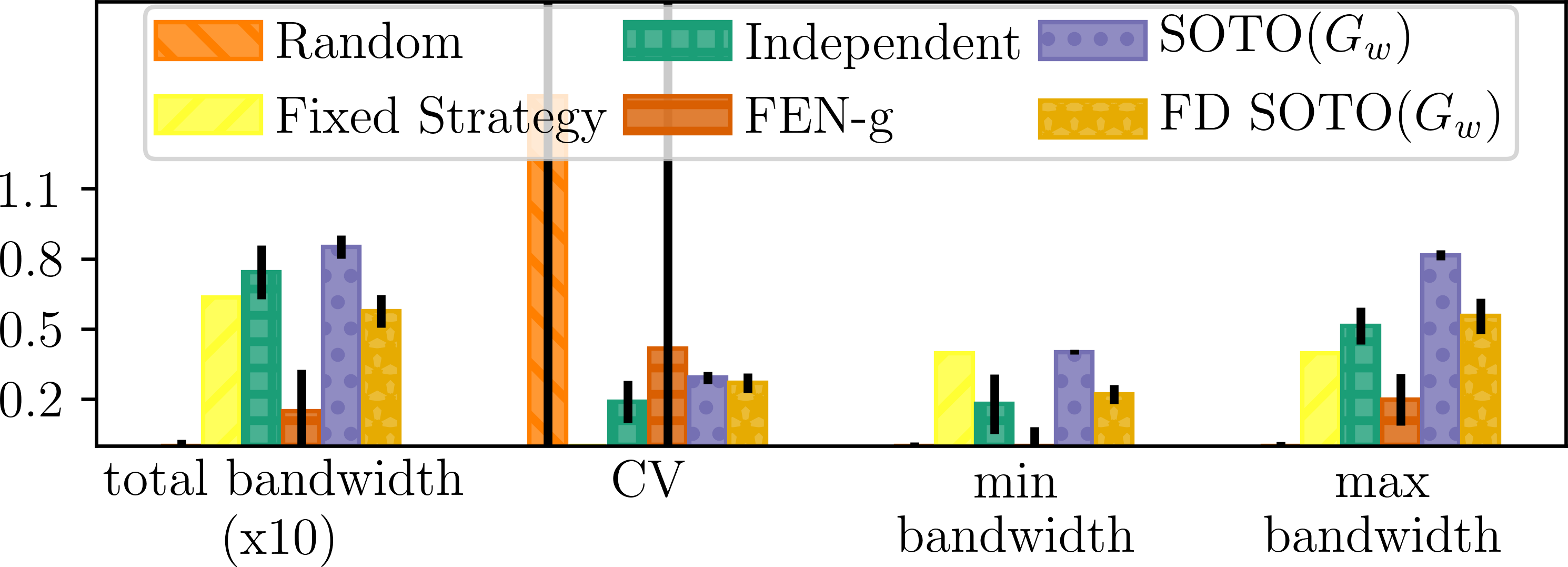}
    \caption{Algorithms' performances in data center control.}
    \label{fig:iroko}
\end{figure}

As \pw{expected}, Figure \ref{fig:centralizedcritic} shows that COMA is better than Independent in terms of total income. 
\us{In Job Scheduling, WQMIX performs \cg{well} in terms of total income but it has the worst CV, while COMA($G_{\bm w}$) and CC($G_{\bm w}$) reach the lowest (better) CV.
It shows that our proposition of optimizing a SWF to achieve fairness can easily be extended with a centralized critic.
Due to the availability of $(\bm o, \bm a)$, \eqref{eq:advapprox} can be computed with less bias, which explains why it can outperform SOTO. However, \cg{due to the} centralized critic and value function, those variants of COMA and vanilla WQMIX are not able to scale in Matthew Effect where they perform poorly compared to SOTO($G_{\bm w}$).}

}

Using the SUMO domain, we further demonstrate that our method can scale up to more complex control tasks, even with unequal access to resources.
For this domain, we added two classic baselines.
At each time step, "Random" selects an action according to a uniform random distribution and "Fixed Strategy" cycles between the traffic-light phases following an optimized period.
Note that $\alpha$-fairness can not be directly applied here because of negative rewards.
Figure~\ref{fig:sumo_perf} shows that in terms of global waiting time, Independent works better than Random, but worse than Fixed Strategy, which means that being too selfish in this domain makes the task harder to solve globally.
\cg{On the contrary,} if the agents cooperate, the traffic flows more smoothly.

Our methods SOTO($G_{\bm w}$) and FD SOTO($G_{\bm w}$) are able to reach the lowest waiting times.
The latter performs \us{better than FEN and WQMIX on} 
all other dimensions (global waiting time, CV, and max waiting time).
FEN achieves a lower CV than Random but at the cost of the worst global waiting time. 
Note that FEN diverges in this environment.

Using the data center control problem, we show how well our method can perform on continuous action spaces. 
To do so, we extended FEN to continuous actions and we also added two classic baselines, "Random" and "Fixed Strategy".
"Random Policy" selects an action according to a uniform random distribution and "Fixed Strategy" always chooses an optimized fixed bandwidth for each host.
Note that $\alpha$-fairness cannot also be applied directly here due to negative rewards.
\mz{In Figure \ref{fig:iroko}, }as expected, the random policy
performs worse as it has the lowest total bandwidth.
The fixed policy performs better than random but worse than RL algorithms except FEN.
Our method with GGF has a lower CV than current state-of-art FEN and the random policy.
The fixed policy has the lowest CV as \mz{the} same action is applied to all agents.
In terms of total bandwidth our method performs very well as it maintains the maximum and minimum bandwidths.

\paragraph{Ablation Study}

\begin{figure}[t]
    \centering
    \includegraphics[width=\linewidth]{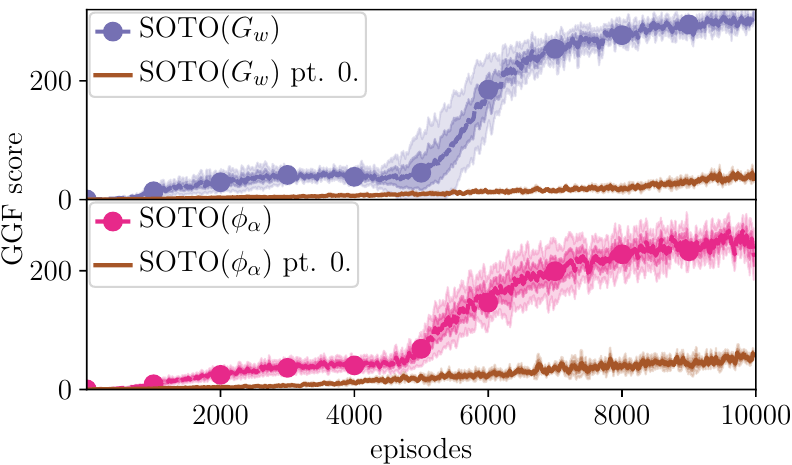}
    \caption{\mz{Comparison of SOTO and SOTO with a randomly initialized \self{} policy on Matthew Effect.}}
    \label{fig:projection}
\end{figure}

\begin{figure}[t]
    \centering
    \includegraphics[width=\linewidth]{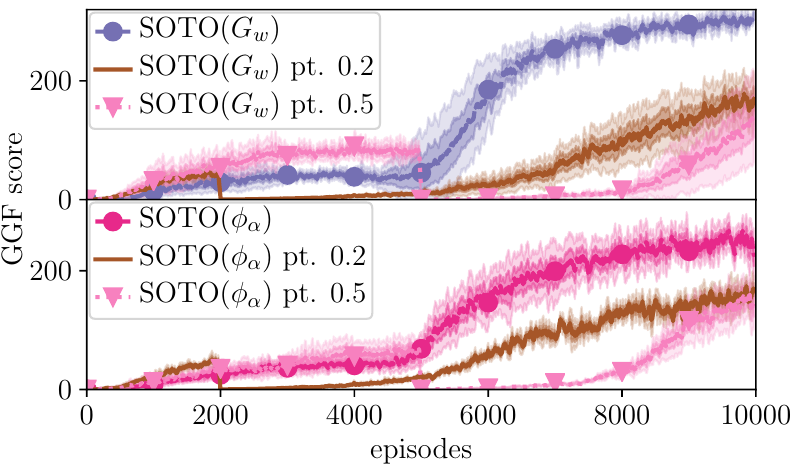}
    \caption{Comparison of SOTO($G_{\bm w}$) and SOTO($\phi_\alpha$) with/without pretraining in Matthew Effect in the CLDE scenario.}
    \label{fig:beta_vs_pretrained}
\end{figure}

\mz{
We first check that the information contained in the \self{} policy is really used by the \team{} policy.
One could argue that SOTO works better than Basic because of the additional inputs (the projection of the observation).
To verify this hypothesis, we trained SOTO with a randomly initialized \self{} policy (equivalent of using the pretraining baseline with $x=0$).
Therefore $\beta$ is not used in this baseline.}
\mz{In Figure \ref{fig:projection}, it is clear that the information gathered in the \self{} policy is important to optimize the SWF.}

To justify the use of $\beta$, we compare our approach in Matthew Effect with two other baselines with pretraining, i.e., the \self{} policy is trained first, then the \team{} one is trained. 
\mz{Those baselines are labeled with the "pt. $x$" tag where $x$ refers to the ratio of the episode dedicated to the pretraining.
Figure \ref{fig:beta_vs_pretrained} clearly \cg{indicates} that training incrementally with $\beta$ by switching the policy used is more data efficient than using pretraining.
}

We refer the reader to Appendix~\ref{app:moreexp} for additional experiments analyzing SOTO, FEN, and Basic.

\section{Conclusion}
\label{sec:conclusion}

We justified and formalized in a theoretically founded way the problem of fair policy optimization in the context of cooperative multi-agent reinforcement learning with independent policies.
We proposed a simple, general and scalable method with a novel neural network architecture allowing an agent to learn to be \cg{first self-concerned} in order to be able to reach a fair solution in a second step.
We furthermore provided a theoretical convergence analysis of policy gradient for this fair optimization problem.
We experimentally shown that each component of our proposed method is useful and that our approach achieves state-of-the-art results on various domains in two different training scenarios.

As future work, the relaxation of impartiality 
or the simultaneous learning of the \self{} and the \team{} policies will be considered.
Another interesting question is how to learn to communicate to achieve fairer solutions.

\section{ Acknowledgments}
This work is supported in part by the program of National Natural Science Foundation of China (No. 61872238), the program of the Shanghai NSF (No. 19ZR1426700), and a Yahoo FREP grant.
Half of the experiments presented in this paper were carried out using the Grid’5000 testbed, supported by a scientific interest group hosted by INRIA and including CNRS, RENATER and several universities as well as other organizations (see \href{https://www.grid5000.fr}{https://www.grid5000.fr}).

\bibliography{manually_added}
\bibliographystyle{icml2021}

\clearpage
\onecolumn
\appendix



\section{Convergence Analysis} \label{app:convergence}

\subsection{Preliminaries}

Let $\mathfrak{J}= \phi \circ \vJ: \mathbb{R}^{N \times K} \rightarrow \mathbb{R}$ denote the \textit{Social Welfare Objective} (SWO), as the composition of the social welfare function $\phi$ with the vectorial objective $\vJ$\paul{, where $N$ is the number of agents and $K$ denotes} the dimension of the parameter $\bm \theta_{i}$ for the policy network of an agent $i \paul{\in [N]}$.

Under a gradient descent optimization scheme, the direction of the update should follow the gradient at \paul{iteration} $k$, \paul{i.e.,}
 \begin{equation} \label{eq:gradientstar}
 \mathfrak{g}^{k, \ast}\coloneqq\nabla\mathfrak{J}(\bm \theta^{k})= \nabla \swf (\vJ(\bm{\theta}^{k}))^\intercal \cdot \nabla \vJ(\bm{\theta}^{k}),
 \end{equation}
 where $ \nabla \vJ$ denotes the Jacobian of $\vJ$.
 \paul{While the subscript on $\bm \theta$ refers to the different agents, superscript $k$ refers to the successive iterations of  policy; it may be omitted below to lighten the notations.}
 
 Adapting the Policy Gradient Theorem \citep{SuttonBarto98} to this fair-multi-agent context, leads us to the following gradient direction for agent $i$: 
 \begin{lemma}
\begin{equation} \label{eq:gstar}
 \nabla_{\bm \theta_{i}}\mathfrak{J}(\bm \theta^{k})
= \beta_{k}  \mathbb{E}_{\bm \theta} \Big[ \bm{A}^{k,\text{SWF}}(\bm s, \bm a) \cdot { \nabla_{\bm \theta_i} \log \pi^{k}_{\bm \theta_i}(a_i|o_i) } \Big],
\end{equation}
where $\bm{A}^{k,\text{SWF}} :=\nabla \swf (\vJ(\bm{\theta}^{k}))^\intercal \cdot  \bm{A}^{k}:  \St \times \Ac \rightarrow \mathbb{R}$, and $\beta_{k}$ is the average length of an episode while following the policy $\bm \pi^{k}$ in the episodic case \paul{or $\beta_k=1$} in the \pw{infinite horizon} case.\footnote{In the infinite horizon case, we can follow a nearly-identical proof, upon the assumption the underlying MDP is ergodic.}
 \end{lemma}
\begin{proof}
Let us look at $\vJ_{(j)}(\bm \theta)\coloneqq\vJ_{I_{j}}(\bm\theta)$, the vectorial objective of agent $i$, and roughly follow the classical policy gradient theorem's  proof. We immediately have:
\begin{equation*}
\nabla_{i} \bm V_{(j)}^{\bm \pi}(s)=  \sum_{a\in \Ac} \left( \nabla_{i}  \pi_{\bm \theta} (\bm a \mid \bm s)    \vQ^{\bm \pi}_{(j)}(\bm s , \bm a) +  \pi_{\bm \theta} (\bm a \mid \bm s) \nabla_{i} \vQ^{\bm \pi}_{(j)}(\bm s , \bm a)  \right)
\end{equation*}
where $\nabla_{i}$, $\vQ_{(j)}$ and $\bm V_{(j)}$ stands for $\nabla_{\bm\theta_{i}}$, $\vQ_{I_{j}}$ and $\bm V_{I_j}$ respectively. 
For the second term, we can develop $\vQ^{\bm \pi}_{(j)}$ as $\sum_{s',r}P(s',r| s,a )(r+\bm V^{\bm \pi}_{(j)}(\bm s') )$. Since nor the reward nor the transition probability depends on $\theta$, it leads to, after simplifications:
\begin{equation}\label{eq:pg1}
\nabla_{i} \bm V_{(j)}^{\bm \pi}(s)= \psi_{i}(\bm s) + \sum_{a\in \Ac} \pi_{\bm \theta} (\bm a | \bm s) \sum_{s'\in \St}   P(\bm s'| \bm s, \bm a ) \nabla_{i} \bm V^{\bm \pi}_{(j)}(\bm s')= \psi_{i}(\bm s) + \sum_{s'\in \St}   \rho^{\bm \pi}(\bm s \rightarrow \bm s', 1) \nabla_{i} \bm V^{\bm \pi}_{(j)}(\bm s') ,
\end{equation}
where  $\psi_{i,j}(\bm s)\coloneqq \sum_{a\in \Ac}  \nabla_{i}  \pi_{\bm \theta} (\bm a| \bm s) \vQ^{\bm \pi}_{(j)}(\bm s , \bm a) $, and $ \rho^{\bm \pi}(\bm s \rightarrow \bm s', k)$ is the probability of transitioning from state $s$ to state $s'$ while following the policy $\bm \pi$ after $k$ steps. Notably $\rho^{\bm \pi}(\bm s \rightarrow \bm s', 1)= \sum_{a\in \Ac} \pi_{\bm \theta} (\bm a | \bm s)  P(\bm s'| \bm s, \bm a )$.
The equation \eqref{eq:pg1} enables us to smoothly unroll a recursive process:
\begin{multline*}
\nabla_{i} \bm V_{(j)}^{\bm \pi}(s)= \psi_{i,j}(\bm s) + \sum_{s'\in \St}   \rho^{\bm \pi}(\bm s \rightarrow \bm s', 1)\left[ \psi_{i,j}(\bm s') + \sum_{\bm s ''\in \St} \rho^{\bm \pi}(\bm s' \rightarrow \bm s'', 1)\nabla_{i} \bm V_{(j)}^{\bm \pi}(s'')  \right] \\
=\psi_{i,j}(\bm s) +  \sum_{s'\in \St}  \rho^{\bm \pi}(\bm s \rightarrow \bm s', 1)\psi_{i,j}(\bm s') + \sum_{s'\in \St}  \sum_{\bm s ''\in \St}  \rho^{\bm \pi}(\bm s \rightarrow \bm s'', 2) \nabla_{i} \bm V_{(j)}^{\bm \pi}(s'') \\
=\cdots \text{ (iterating) } \cdots \\
= \sum_{\bm s' \in \St }\sum_{k=0}^{\infty} \rho^{\bm \pi}(\bm s \rightarrow \bm  s', k) \psi_{i,j}(\bm  s'),
\end{multline*}
Coming back to the derivative of the vectorial objective of agent $j$:
\begin{equation*} 
 \nabla_{i}\vJ_{(j)}(\bm \theta)
= \sum_{\bm s \in \St } \chi^{\bm \pi}(\bm s) \psi_{i,j}(\bm  s)=  \beta^{\bm \pi}  \left( \sum_{\bm s \in \St } d^{\bm \pi}(\bm s)\psi_{i,j}(\bm  s) \right) ,
\end{equation*}
where $\chi^{\bm \pi}(\bm s)\coloneqq \sum_{k=0}^{\infty} \rho^{\bm \pi}(\bm s_{0} \rightarrow \bm  s, k) $, $d^{\bm \pi}(\bm s)=\frac{\chi^{\bm \pi}(\bm s)}{\sum_{\bm s'\in \St} \chi^{\bm \pi}(\bm s')}$ is the stationary distribution and $\beta^{\bm \pi} :=\sum_{\bm s \in \St } \chi^{\bm \pi}(\bm s)$.
Since $ \pi_{\bm \theta} (\bm a| \bm s) =\sum_{\bm o } \Omega( \bm o | \bm s) \prod_{i}  \pi_{\bm \theta_{i}} ( a_{i}| o_{i}) $:
\begin{multline*}
 \nabla_{i}\vJ_{(j)}(\bm \theta)=\beta^{\bm \pi} \sum_{\bm s\in \St}   \sum_{\bm a\in \Ac} d^{\bm \pi}(\bm s) \vQ^{\bm \pi}_{(j)}(\bm s , \bm a) \nabla_{i}  \pi_{\bm \theta} (\bm a| \bm s) = \beta^{\bm \pi} \sum_{\bm s\in \St}\sum_{a\in \Ac} d^{\bm \pi}(\bm s)\pi_{\bm \theta} (\bm a| \bm s)  \vQ^{\bm \pi}_{(j)}(\bm s , \bm a)   \nabla_{i}  \log \pi_{\bm \theta} (\bm a| \bm s)\\
 =  \beta^{\bm \pi} \mathbb{E}_{
   {  {\bm s \sim d^{\bm \pi}(\bullet ) \atop
     \bm a  \sim \pi (\bullet| \bm s) } \atop
     \bm o \sim \Omega(\bullet | \bm s)}
} \Big[ \vQ^{\bm \pi}_{(j)}(\bm s , \bm a) \nabla_{\bm \theta_i} \log \pi_{\bm \theta_i}(a_i|o_i) \Big]
\end{multline*}
These computations may be equivalently rewritten with the advantage function: 
\begin{equation*}
 \nabla_{i}\vJ (\bm \theta) = \beta^{\bm \pi}  \mathbb{E} \Big[ \bm A^{\bm \pi}(\bm s , \bm a) \nabla_{\bm \theta_i} \log \pi_{\bm \theta_i}(a_i|o_i) \Big]
\end{equation*}
Hence, returning to the SWO $\mathfrak{J}$, it boils down to the Lemma's claim:
\begin{equation*}
 \nabla_{i}\mathfrak{J} (\bm \theta) = \beta^{\bm \pi} \nabla \swf(\vJ(\bm \theta))^{T} \cdot \mathbb{E} \Big[ A^{\bm \pi}(\bm s , \bm a)\nabla_{\bm \theta_i} \log \pi_{\bm \theta_i}(a_i|o_i) \Big]=  \beta^{\bm \pi} \mathbb{E} \Big[ A^{\bm \pi, SWF}(\bm s , \bm a)\nabla_{\bm \theta_i} \log \pi_{\bm \theta_i}(a_i|o_i) \Big]
\end{equation*}
\end{proof}

As mentioned previously, since we adopt a fully decentralised framework with partial observability and since the advantage function is not known, we instead  \paul{approximate} our gradient update \paul{according to the following direction} $\tilde{\mathfrak{g}}_{i}^{k}$ for an agent $i$: 
\begin{equation*} 
\tilde{\mathfrak{g}}_{i}^{k}\coloneqq \mathbb{E}_{\bm \theta} \Big[ \bm{\hat{A}}^{k,\text{SWF}}(\bm o, \bm a) \cdot { \nabla_{\bm \theta_i} \log \pi_{\bm \theta_i}(a_i|o_i) } \Big],
\end{equation*} 
where $\bm{\hat{A}}^{k}= (\bm{\hat{A}}_{i}^{k})_{i \in [N]}$ aggregates the estimate of the local advantage at step \paul{$k$} for each agent \paul{$i$}, $\bm{\hat{A}}_{i}^{k}: \Ob_i \times \Ac_i \rightarrow \mathbb{R}^{\mid I_i \mid}$, and $\bm{\hat{A}}^{k,\text{SWF}}(\bm o, \bm a)=\nabla \swf (\hat{\vJ} (\bm{\theta}^{k}))^\intercal \cdot  \hat{\bm{A}}^{k}(\bm o, \bm a)$. 

Let $\hat{\mathfrak{g}}^{k} \in \mathbb{R}^{N\times K}$ \paul{denote the estimated gradient} at step \paul{$k$}, estimated from sampling a minibatch of transitions of size $m$.
The update at step $k$ of this stochastic gradient ascent can be written as:
\begin{equation} \label{eq:updatetheta}
\bm \theta^{k+1}\leftarrow \bm \theta^{k} +\alpha_{k} \hat{\mathfrak{g}}^{k}
\end{equation}

Comparing the ideal and estimated gradients by introducing the error term $\err^{k}$, using \eqref{eq:gstar}:
\begin{equation}\label{eq:err}
    \hat{\gr}^{k} =  \overline{\gr}^{k} + \err^{k}= \beta_{k}\gr^{k, \ast} + \err^{k},  \quad\text{ where } \quad \overline{\gr}_{i}^{k} :=\mathbb{E}_{\bm \theta} \Big[ \bm{A}^{k,\text{SWF}}(\bm s, \bm a) \cdot { \nabla_{\bm \theta_i} \log \pi_{\bm \theta_i}(a_i|o_i) } \Big].
\end{equation}
Let us decompose this error term into two parts:
\begin{equation}\label{eq:err2}
 \err^{k}= \bm{\nu}^{k}+\bm{\eta}^{k} , \quad \text{ where } \bm{\nu}^{k}\coloneqq \hat{\gr}^{k}- \tilde{\gr}^{k} \quad \text{ and } \quad \bm{\eta}^{k} \coloneqq\tilde{\gr}^{k} - \overline{\gr}^{k}
\end{equation}
While the first term $\bm{\nu}^{k}$ tends towards zero when $m$ tends towards infinity, some assumptions would be made below to control the second part of this error $\bm{\eta}^{k}$.
 
\subsection{Assumptions}

\paul{Here we detail and discuss the assumptions made in our theoretical analysis.}
The first part of \paul{the} assumptions regarding the social welfare function $\swf$ \paul{has} already been justified in the context of fairness considerations in Section~\ref{sec:fairness}. 
We simply assume \paul{in addition} that the \paul{norm of the} gradient of $\phi$ is bounded by $M_{\swf} \paul{\in \mathbb R^+}$ on the image of $\vJ$.
\begin{assumption} \label{as:hypswf}
 $\swf$ is concave and non-decreasing in each argument. 
 It is (sub-)differentiable\footnote{We could adapt the proof to the case of sub-differentiability of $\phi$ and $\vJ$, yet their sub-gradient should satisfy Lipschitz-inequalities.} and \paul{the norm of} its gradient is assumed \paul{to be} bounded by $M_{\swf} \paul{\in \mathbb R^+}$ on the image of $\vJ$. 
 \paul{This} implies that $\swf$ is $L_{\swf}$-smooth on the image of $\vJ$, with $L_{\swf} \leq M_{\swf}$. 
\end{assumption}

Adopting the boundedness of the reward function is a common and legitimate hypothesis in the literature of policy gradient or actor-critic algorithms. If $B_{R}$ is uniformly bounding the reward, in the discounted case, $\frac{B_{R}}{1-\gamma}$ would uniformly bound both the value or action-value function, and $\frac{2B_{R}}{1-\gamma}$ would bound the advantage function. From these considerations, it seems legitimate to assume the estimated advantage function \paul{to be} bounded:
\begin{assumption} \label{as:hypAbound}
The estimated advantage function $\hat{A}^{k}$ is uniformly bounded by $B_{A}$, for all k.
\end{assumption}

Here, we adopt classic assumptions concerning the vectorial objective function $\vJ$:
\begin{assumption} \label{as:hypJ}
The function $\vJ: \mathbb{R}^{N \times K} \rightarrow \mathbb{R}^{D}$ is concave and Lipschitz-smooth with constant $L_{\vJ}$, i.e., its gradient is $L_{\vJ}$-Lipschitz continuous. 
These two hypothes\paul{es} imply respectively:
 \begin{equation*}
 \forall i \in [D], \quad   \vJ_{i}(\bm \theta')- \vJ_{i}(\bm \theta) \leq \langle \nabla  \vJ_{i}(\bm \theta) , \bm \theta'-\bm \theta\rangle 
 \end{equation*}
 \begin{equation*}
  \norm{ \nabla \vJ(\bm \theta')- \nabla \vJ(\bm \theta)} \leq L_{\vJ} \norm {\bm \theta'-\bm \theta} 
 \end{equation*}
\end{assumption}
\pw{This assumption notably covers the case where a linear approximation scheme is used.}
 By Cauchy-Schwartz, integrating the last equation, and using the concavity hypothesis:
 \begin{equation}\label{eq:hypJ2}
   \vJ(\bm \theta')- \vJ(\bm \theta) \geq \nabla \vJ(\bm \theta)^\intercal (\bm \theta'-\bm \theta) - \frac{L_{\vJ}}{2} \norm {\bm \theta'-\bm \theta}^{2}
 \end{equation}

\pw{The previous assumptions imply:}
\begin{lemma}\label{jswfconcave}
\begin{enumerate}
    \item[$(i)$] Under Assumptions  (\ref{as:hypswf}, \ref{as:hypJ}), the SWO $\mathfrak{J}$ is concave. 
    \item[$(ii)$] Under Assumptions (\ref{as:hypswf}, \ref{as:hypJ}), with the additional hypothesis that the gradient of the vectorial objective $\nabla \vJ$ is bounded by $M_{\vJ}$, then the SWO $\mathfrak{J}$ is $L_{\mathfrak{J}}-$smooth, with:
$$L_{\mathfrak{J}}\leq L_{\swf}M_{\vJ}^{2} + M_{\swf} L_{\vJ}.$$
\end{enumerate}
\end{lemma}
\begin{proof}
\begin{enumerate}
    \item[$(i)$]
From Assumptions \ref{as:hypswf}  and \ref{as:hypJ},  both $\swf$ and $\vJ$ are concave, and $\swf$ is non-decreasing in each argument. It is then straightforward to prove that the fair objective $\mathfrak{J}=\swf \circ J$ is concave.
    \item[$(ii)$]
    For the last assertion, we can \paul{show}:
\begin{multline*}
    \norm{\nabla \mathfrak{J}(\bm \theta')- \nabla \mathfrak{J}(\bm \theta)}= \norm{\nabla \swf(\vJ(\bm \theta'))^\intercal\nabla\vJ(\bm \theta')-\nabla \swf(\vJ(\bm \theta))^\intercal\nabla\vJ(\bm \theta) }\\
    \leq \norm{\nabla \swf(\vJ(\bm \theta'))^\intercal\nabla\vJ(\bm \theta')-\nabla \swf(\vJ(\bm \theta))^\intercal\nabla\vJ(\bm \theta') } + \norm{\nabla \swf(\vJ(\bm \theta))^\intercal\nabla\vJ(\bm \theta')-\nabla \swf(\vJ(\bm \theta))^\intercal\nabla\vJ(\bm \theta)}\\
    \leq L_{\swf} \norm{\vJ(\bm \theta')-\vJ(\bm \theta)} M_{J} + L_{J}M_{\swf} \norm{\bm \theta'-\bm \theta}\leq \left( L_{\swf}M_{J}^{2} + L_{J}M_{\swf}\right) \norm{\bm \theta'-\bm \theta}.
\end{multline*}
\end{enumerate}
\end{proof}

Let $\Theta^{\ast}$ denote the set of \pw{maximizers of} $ \mathfrak{J}$, i.e., $\Theta^{\ast}\coloneqq \argmax \{ \mathfrak{J}(\bm \theta) \mid \bm \theta\in \mathbb{R}^{N \times K} \}$. 
The concavity of $ \mathfrak{J}$ implies that the set $\bm \theta^{\ast}$ is convex. We furthermore assume:
\begin{assumption} \label{as:hypR}
The optimal set $\Theta^{\ast}$ is non empty and there exists a radius bound $R_{0}$ such that:
$$\max_{\bm \theta^{\ast}\in  \Theta^{\ast}} \max_{\bm \theta}   \lbrace \norm{\bm \theta-\bm \theta^{\ast}} \mid  \mathfrak{J}(\bm \theta) \geq    \mathfrak{J}(\bm \theta^{0}) \rbrace \leq R_{0},$$
\paul{where} $\theta^{0}$ is assumed to be the initial parameter value.%
\footnote{However, if such hypothesis is not satisfied for $\bm \theta^{0}$, we may assume it is satisfied for another $\bm \theta^{i_{0}}$; it alters the inequality formulation in our theorem ---as we need to shift the indices by $i_{0}$---, but not the convergence result itself.}
\end{assumption}


A common condition on the learning rate in convex optimization is to assume it is smaller than the inverse of the Lipschitz constant. 
Here, we use a slight variant of this condition: 
\begin{assumption}\label{as:hypalpha}
Given Lipschitz constant $L_{\vJ}$ from  Assumption \ref{as:hypJ}, and gradient bound $M_{\swf}$ from  Assumption \ref{as:hypswf}, the learning rate satisfies:
$$\alpha_{k} \leq \frac{1}{L_{\vJ} M_{\swf} \beta_{k}},$$
where $\beta_{k}$ is the average length of an episode while following the policy $\bm \pi^{k}$ in the episodic case \paul{or $\beta_k = 1$} in the \pw{infinite horizon} case.
\end{assumption}

Adopting classic assumptions regarding the policy, in the policy gradient or actor-critic literature of convergence analysis:%
\footnote{Assumption $5.1$ in \cite{pmlr-v115-xu20a}, Assumption $3.1$ in \cite{zhang_global_2020}, Assumption $4.1$ in \cite{pmlr-v80-papini18a}, Assumption $1$ in \cite{kumar_sample_2019}, Assumption $4.5$ in \cite{qiu_finite-time_nodate}, etc. 
}
\begin{assumption} \label{as:hyppi}
\begin{enumerate}
    \item[$(i)$] The policy $\bm \pi_{\bm \theta}$ is differentiable with respect to $\bm \theta$, and $\nabla \log \bm \pi_{\bm \theta}(\bm a | \bm s)$, referred to as the score function, exists. Moreover, there exists a bound $M_{\bm \pi}\in\mathbb{R}^{+}$ which uniformly bounds $\norm{\nabla \log \bm \pi_{\bm \theta}}$ for all $\bm \theta$:
$$\sup_{\bm \theta, \bm s, \bm a} \norm{\nabla \log \bm \pi_{\bm \theta}(\bm a | \bm s)} = M_{\bm \pi}$$
\item[$(ii)$] The variance of the estimated gradient is uniformuly bounded:
$$\sup_{\bm \theta, \bm s, \bm a} \Var(\hat{\gr}^{\bm \theta}(\bm s, \bm a)) \leq \sigma^{\musSharp{}}$$
\end{enumerate}
\end{assumption}

In our context of fully decentralized and partially observable multi-agent RL, we have to consider the error made by the local estimates of the advantage function, as mentioned in \eqref{eq:advapprox}.
Within such an approximation lies several sources of error that we will consider separately \paul{below}.

First, we need to approximate the objective $\vJ$.
\pw{Since we want to focus our analysis on the convergence of the \paul{policy}%
\footnote{In Actor-Critic schemes, it is common to adopt a two-timescale analysis, in which we usually need to ensure that the actor update moves slowly while the critic update chases the slowly-moving target defined by the actor by enforcing a condition on the ratio of their learning rates (cf. \cite{NIPS1999_6449f44a}). 
Here, we do not focus on bounding the error made from the advantage estimation in probability, in function of the size of the mini-batch $m$, but the analysis could be extended in later work.}, we make the following assumption\paul{s}:}
\begin{assumption} \label{as:hypJhat}
The approximation error of $\vJ$ is bounded in probability:%
\footnote{If we adopt GGF as SWF, this hypothesis could be relaxed, as we would only care that the estimate of \paul{$\vJ$} preserves the order of the different $J_{i}(\bm \theta^{k})$ for each agent, not the actual value.
}
$$\forall k>0, \quad \prob\left(\norm{ \hat{\vJ}(\bm \theta^{k})-\vJ(\bm \theta^{k})} \leq \Xi_{J}\right) =1 $$
\end{assumption}

Then, we have to control the approximation error of the local advantage function, which stems from several sources (such as the finiteness of the sample set on which it would be estimated, and the expressiveness of the class of functions estimators used for $\hat{Q}$ and $\hat{V}$):
\begin{assumption} \label{as:hypAhat}
The approximation error of the local advantage $\bm A$ is uniformly bounded in probability:
$$\forall k>0, \quad \prob\left(\norm{ \hat{\bm{A}}^{k}(\bm o, \bm a)-\bm{A}^{k}(\bm o, \bm a)}_{\infty} \leq \Xi_{A}\right)=1$$
\end{assumption}

Finally, the mismatch between the local and global advantage functions has to be taken into consideration:
\begin{assumption} \label{as:hypAinter}
The error stemming from the use of the decentralized local advantage is uniformly bounded in probability:
\begin{equation*}
\forall k>0, \forall i\in [N], \forall j\in [N], j \neq i,  \quad
\prob\left(\norm{ \bm{A}_{j/i}^{k}(o_{i},a_{i})}_{\infty} \leq \Xi_{L}\right) =1
\end{equation*}
Or, less restrictively, after application of the SWF $\swf$, we can assume instead:
\begin{equation}\label{eq:AjiSWF}
\forall k>0, \forall i\in [N], \quad
\prob\left(\norm{ \bm{A}_{\bullet /i}^{(-i), {k,\text{SWF}}}(o_{i},a_{i})}_{\infty} \leq \Xi_{L}\right)=1,
\end{equation}
$$\text{ with } \bm{A}_{j/i}^{k}(o_{i},a_{i})\coloneqq \mathbb{E}_{\bm o_{-i}\sim \prob(\cdot \mid o_{i}), \atop \bm a_{-i} \sim \bm \pi^{-i}(\cdot \mid \bm o_{-i})}\Big[ \bm{A}_{j}^{k}(\lbrace o_{i},\bm o_{-i} \rbrace ,\lbrace a_{i},\bm a_{-i}\rbrace ) \Big] \text{ and }  \left(\bm{A}^{(-i), k}_{\bullet/i} (o_{i},a_{i})\right)_{j}\coloneqq \left\{\begin{array}{cc}
    \bm{A}^{k}_{j/i} (o_{i},a_{i})  & j \neq i \\
    0  & j=i \end{array} \right. .$$
\end{assumption}
Here, the notations $\bm o_{-i}$ resp. $\bm a_{-i}$ designate the set  $\lbrace o_{j}\rbrace_{j\neq i}$ resp. $\lbrace a_{j}\rbrace_{j\neq i}$.
The notation $\bm{A}_{j/i}$ stands for the advantage of the actions of agent $i$ relative to another agent $j$.
As made clear in the proof of Lemma~\ref{lemmaeta}, when comparing $\bm{A}(\bm s,\bm a)$ and $\bm{A}(\bm o,\bm a)$, these inter-agent advantages $\bm{A}^{(-i)}_{\bullet/i}$ appear. This assumption tackles the discrepancy between each agent\paul{'s} local advantage and each agent\paul{'s} global advantage, which would take into consideration the benefit of the actions of other agents. 
Looking at \eqref{eq:AjiSWF}, let us remark that the benefits (positive or negative) of one agent action, may, upon application of the SWF $\phi$, average out in some situations. 
\paul{Therefore, ignoring them may not have too much impact.}

Note that the conditional probability arising in the above expected value satisfies, with the previous notations:
\begin{equation*}
\prob(\bm o_{-i} \mid o_{i}) 
= \sum_{\bm s} d^{\bm \pi}(\bm s) \prob (\bm o_{-i} \mid o_{i}, \bm s) 
= \sum_{\bm s} \frac{ d^{\bm \pi}(\bm s) \Omega (\bm o \mid \bm s) }{\prob(o_{i} \mid \bm s)} 
= \sum_{\bm s} \frac{ d^{\bm \pi}(\bm s) \Omega (\bm o \mid \bm s) }{\Omega (o_{i} \mid \bm s)} 
\end{equation*}

These assumptions enable to bound the error term $\bm{\eta}^{k}$ as defined in \eqref{eq:err2}: 
\begin{lemma}\label{lemmaeta}
Under assumptions \ref{as:hypswf},\ref{as:hypAbound},\ref{as:hyppi},\ref{as:hypJhat},\ref{as:hypAhat}, \ref{as:hypAinter}, $\bm\eta^{k}$ is bounded in probability:
$$\forall k\quad \prob\left(\norm{\bm{\eta}^{k}} \leq \Xi \right)=1,\quad  \text{  with   } \quad \Xi=(L_{\swf} B_{A} \Xi_{J} + M_{\swf} \Xi_{A} + M_{\swf}  \Xi_{L}) M_{\bm \pi},$$
with the bounds as previously defined. 
\end{lemma}
\begin{proof}
Splitting into three terms, from the definition of $\bm{A}^{SWF}$ :
\begin{align} \label{eq:threeterms}
\norm{\bm{\eta}_{i}^{k}} =\norm{ \tilde{\gr}_{i}^{k} - \overline{\gr}_{i}^{k}} = & \norm{  \mathbb{E}_{\bm \theta} \Big[ \left( \bm{\hat{A}}^{k,\text{SWF}}(\bm o, \bm a) - \bm{A}^{k,\text{SWF}}(\bm s, \bm a)\right)  { \nabla_{\bm \theta_i} \log \pi^{k}_{\bm \theta_i}(a_i|o_i) } \Big] } \notag\\
    \leq & \norm {  \mathbb{E}_{\bm \theta} \Big[ \left(\nabla\swf(\hat{\vJ}(\bm \theta^{k}))- \nabla\swf\left(\vJ(\bm \theta^{k})\right)\right)^\intercal \cdot   \bm{\hat{A}}^\text{k}(\bm o, \bm a)   \nabla_{\bm \theta_i} \log \pi^{k}_{\bm \theta_i}(a_i|o_i) \Big] } \notag\\
    & + \norm{  \mathbb{E}_{\bm \theta} \Big[\nabla\swf\left(\vJ(\bm \theta^{k})\right)^\intercal \cdot ( \bm{\hat{A}}^\text{k}(\bm o, \bm a) - \bm{A}^\text{k}(\bm o, \bm a))  \nabla_{\bm \theta_i} \log \pi^{k}_{\bm \theta_i}(a_i|o_i) \Big] }
    \notag\\
    & + \norm{  \mathbb{E}_{\bm \theta} \Big[ \nabla\swf\left(\vJ(\bm \theta^{k})\right)^\intercal \cdot ( \bm{A}^\text{k}(\bm o, \bm a) - \bm{A}^\text{k}(\bm s, \bm a))  \nabla_{\bm \theta_i} \log \pi^{k}_{\bm \theta_i}(a_i|o_i) \Big] }
\end{align}
Since $\norm{\mathbb E(X)} \leq \mathbb E(\norm{X})$:
\begin{align}
\norm{\bm{\eta}_{i}^{k}} \leq & \Big(
L_{\swf} B_{A}\norm{\vJ(\bm \theta^{k})-\vJ(\bm \theta)}
    + M_{\swf} \Xi_{A} + M_{\swf}  \Xi_{L}  \Big) \mathbb{E}_{\bm \theta} \Big[\norm{ \nabla_{\bm \theta_i} \log \pi^{k}_{\bm \theta_i}(a_i|o_i)} \Big] \notag \\
    \leq & \Big(
L_{\swf} B_{A} \Xi_{J}   
    + M_{\swf} \Xi_{A} + M_{\swf}  \Xi_{L}  \Big)M_{\bm \pi}, \label{eq:etabound}
\end{align}
with the previously defined bounds; for the first inequality we used the fact that $\swf$ is $L_{\swf}$-smooth, and Assumptions \ref{as:hypAbound} resp. \ref{as:hypAhat} resp. \ref{as:hypAinter} for the different terms; for the second inequality, we used Assumption \ref{as:hypJhat} and the boundedness of $\norm{\nabla \ln \bm \pi}$ from Assumption \ref{as:hyppi}.\\
Let us justify the computations for the third term of \eqref{eq:threeterms}. 
On the one hand, from the definition of $\bm{A}_{j/i}$, after some simplifications, we arrive at:
\begin{equation*}
 \mathbb{E}_{\bm \theta}\Big[ \bm{A}_{j}^\text{k}(\bm s, \bm a) \nabla_{\bm \theta_i} \log \pi^{k}_{\bm \theta_i}(a_i|o_i) \Big]= \mathbb{E}_{\bm  \theta}\Big[ \bm{A}_{j/i}^\text{k}(o_{i},a_{i}) \nabla_{\bm \theta_i} \log \pi^{k}_{\bm \theta_i}(a_i|o_i) \Big] 
\end{equation*}
On the other hand, the other terms may be simplified to:
\begin{equation*}
 \mathbb{E}_{\bm \theta}\Big[ \bm{A}_{j}^\text{k}(o_{j},a_{j}) \nabla_{\bm \theta_i} \log \pi^{k}_{\bm \theta_i}(a_i|o_i) \Big]= \left\{ \begin{array}{ll}
    0 & j \neq i  \\
     \mathbb{E}_{\bm \theta}\Big[ \bm{A}_{i}^\text{k}(o_{i},a_{i}) \nabla_{\bm \theta_i} \log \pi^{k}_{\bm \theta_i}(a_i|o_i) \Big] & j=i  
 \end{array}\right.
  \end{equation*}

 Therefore, the third term of the right side in \eqref{eq:threeterms}, boils down to:
 \begin{equation*}
     \norm{  \mathbb{E}_{\bm \theta} \Big[ \nabla\swf\left(\vJ(\bm \theta^{k})\right)^\intercal \cdot (\bm{A}^{(-i), k}_{\bullet/i}(o_{i},a_{i}))  \nabla_{\bm \theta_i} \log \pi^{k}_{\bm \theta_i}(a_i|o_i) \Big] },
 \end{equation*}
 where $\bm{A}^{(-i), k}_{\bullet/i}(o_{i},a_{i})$ denotes the vector whose $j$-th component is:  $\left(\bm{A}^{(-i), k}_{\bullet/i} (o_{i},a_{i})\right)_{j}\coloneqq \left\{\begin{array}{cc}
    \bm{A}^{k}_{j/i} (o_{i},a_{i})  & j \neq i \\
    0  & j=i
 \end{array}\right. .$
Assumption \ref{as:hypAinter} enable us to conclude.

\end{proof}

\paragraph{Locally concave Extension}
To extend our analysis to the locally concave scenario, two previous assumptions (Assumptions \ref{as:hypJ}, \ref{as:hypR}) need to be reformulated as follows:
\begin{assumptionbis}[$3'$] \label{as:hypJbis}
There exists a neighborhood $\mathcal{U}\subset \mathbb{R}^{N \times K}$ containing \paul{the initial parameter value} $\theta^{0}$, such that:
\begin{itemize}
    \item The function $\vJ: \mathbb{R}^{N \times K} \rightarrow \mathbb{R}^{D}$ is concave on $\mathcal{U}$.
    It is Lipschitz-smooth on $\mathcal{U}$ with constant $L_{\vJ}$, i.e., its gradient is $L_{\vJ}$-Lipschitz continuous on $\mathcal{U}$.
    \item  The set of optimal values in $\mathcal{U}$ $\Theta^{\mathcal{U},\ast}$ is non empty. There exists a radius bound $R^{\mathcal{U}}_{0}$ such that:
$$ \max_{\theta^{\ast}\in  \Theta^{\mathcal{U},\ast} } \max_{\theta\in \mathcal{U}}   \lbrace \norm{\theta-\theta^{\ast}} \mid  \mathfrak{J}(\theta) \geq    \mathfrak{J}(\theta^{0}) \rbrace \leq R^{\mathcal{U}}_{0}.$$
\end{itemize}
\end{assumptionbis}
\pw{This scenario covers the more general case where a neural network is used for the approximation.}
As previously, together with Assumption (\ref{as:hypswf}), ($i$) implies the local concavity of $\mathfrak{J}$:
\begin{lemma}\label{jswflocconcave}
Under Assumptions (\ref{as:hypswf}, $3'$), the SWO $\mathfrak{J}$ is locally concave. 
\end{lemma}

\subsection{Suboptimality Analysis}

\begin{lemma}\label{lemma1}
Under Assumptions (\ref{as:hypswf}, \ref{as:hypJ}, \ref{as:hypalpha}) the SWO between two consecutive update satisfies:
\begin{equation}\label{eq:eq5}
 \forall k>0, \quad    \mathfrak{J}(\bm \theta^{k+1})-\mathfrak{J}(\bm \theta^{k}) \geq\gamma_{k}
    \norm{\nabla \mathfrak{J}(\bm \theta^{k})}^{2} - \delta_{k}  \norm{\err^{k}}^{2}, \quad \text{ with }\gamma_{k}=\frac{\alpha_{k}\beta_{k}}{2}, \text{ and } \delta_{k}=\frac{\alpha_{k}}{2 \beta_{k}}.
\end{equation}
\end{lemma}

\begin{proof}
Since $\swf$ is assumed concave (Assumption \ref{as:hypswf}):
\begin{equation}\label{eq:eq1}
    \mathfrak{J}(\bm \theta^{k+1})-\mathfrak{J}(\bm \theta^{k}) \geq \nabla \swf (\vJ(\bm \theta^k))^\intercal (\vJ(\bm \theta^{k+1})-\vJ(\bm \theta^{k}) ).
\end{equation}
Since $\vJ$ is assumed $L_{\vJ}-$smooth and concave (Assumption \ref{as:hypJ}), the following inequality holds:
\begin{equation}\label{eq:eq2}
 \vJ(\bm \theta^{k+1})-\vJ(\bm \theta^{k}) \geq \nabla \vJ(\bm \theta^{k})^\intercal (\bm \theta^{k+1}-\bm \theta^{k}) - \frac{L_{\vJ}}{2} \norm{\bm \theta^{k+1}-\bm \theta^{k}}^{2}
\end{equation}
Since $\swf$ is assumed non decreasing in each component (Assumption \ref{as:hypswf}), we can easily combine \eqref{eq:eq1} and \eqref{eq:eq2} to deduce:
\begin{equation}\label{eq:eq3}
    \mathfrak{J}(\bm \theta^{k+1})-\mathfrak{J}(\bm \theta^{k}) \geq \nabla \swf (\vJ(\bm \theta^k))^\intercal (\nabla \vJ(\bm \theta^{k})) (\bm \theta^{k+1}-\bm \theta^{k})^\intercal -  \norm{\swf (\vJ(\bm \theta^k))}_{1} \frac{L_{\vJ}}{2} \norm{\bm \theta^{k+1}-\bm \theta^{k}}^{2}
\end{equation}

Using Assumption \ref{as:hypswf} to bound \paul{the gradient of $\swf$}, and given the definition of $\gr^{k, \star}$ \eqref{eq:gstar}, and the update rule \eqref{eq:updatetheta} the above inequality turns into:
\begin{equation}\label{eq:eq4}
    \mathfrak{J}(\bm \theta^{k+1})-\mathfrak{J}(\bm \theta^{k}) \geq \alpha_{k}
    \langle \gr^{k, \star}, \hat{\gr}^{k} \rangle - \frac{L_{\vJ}M_{\swf}}{2} \alpha_{k}^{2} \norm{\hat{\gr}^{k}}^{2}
\end{equation}

Rewriting \eqref{eq:eq4} with the error in \eqref{eq:err}, and using the condition on the learning rate (Assumption \ref{as:hypalpha}) would provide the result, since:
\begin{align*}
    \alpha_{k}
    \langle \gr^{k, \star}, \hat{\gr}^{k} \rangle - \frac{L_{\vJ} M_{\swf}}{2} \alpha_{k}^{2} \norm{\hat{\gr}^{k}}^{2} \geq & \alpha_{k}
    \langle \gr^{k, \star}, \hat{\gr}^{k} \rangle - \delta_{k}  \norm{\hat{\gr}^{k}}^{2}\\
   = & (\alpha_{k}\beta_{k} -\delta_{k} (\beta_{k})^{2})  \langle \gr^{k, \star}, \gr^{k, \star} \rangle + (\alpha_{k}-2\delta_{k}\beta_{k})  \langle \gr^{k, \star}, \err^{k} \rangle- \delta_{k}\norm{\err^{k}}^{2}\\
   = & \gamma_{k}\langle \gr^{k, \star}, \gr^{k, \star} \rangle - \delta_{k}\norm{\err^{k}}^{2}, 
\end{align*}
    where we set $\delta_{k}=\frac{\alpha_{k}}{2\beta_{k}}$, $\gamma_{k}=\frac{\alpha_{k}\beta_{k}}{2}$. The first inequality comes from Assumption \ref{as:hypalpha}, as it implies $\frac{L_{\vJ} M_{\swf}}{2} \alpha_{k}^{2} \leq \delta_{k}$. As introduced previously, let us remind the constant $\beta_{k}$ is the one stemming from the Policy Gradient Theorem, and corresponds in the episodic case to the average length of an episode.
\end{proof}

Before stating the main theorem, let us invoke a simple general lemma needed for evaluating the rate of convergence, in plain form and in probability.
\begin{lemma}\label{lemma2}
\begin{enumerate}
    \item[$(i)$] If a sequence $u_{k}$ satisfies an equation of the form:
\begin{equation} \label{eq:assumptionlemma2}
\forall k \leq k_{0}, \quad u_{k}-u_{k+1} \geq \mu u_{k}^{2}, \quad \text{with } \mu\in \mathbb{R}^{+}, u_{0}>0,
\end{equation}
Then, it has a sublinear rate of convergence:
\begin{equation*}
\forall k \leq k_{0}, \quad u_{k} \leq \frac{u_{0}}{(\mu u_{0}) k + 1} 
\end{equation*}
    \item[$(ii)$] If a sequence $u_{k}$ satisfies:
\begin{equation} \label{eq:assumptionlemma2proba}
\forall k \leq k_{0}, \quad \prob \left( u_{k}-u_{k+1} \geq \mu u_{k}^{2} \right) > 1- \iota, \quad \text{with }  \mu\in \mathbb{R}^{+},  u_{0}>0.
\end{equation}
Then, it has a sublinear rate of convergence with a certain probability:
\begin{equation*}
\forall k \leq k_{0}, \quad \prob \left( u_{k} \leq \frac{ u_{0}}{(\mu  u_{0})k+1} \right) \geq (1- \iota)^{k}
\end{equation*}
\end{enumerate}
\end{lemma}
\begin{proof}
\begin{enumerate}
    \item[$(i)$] First, let us remark that the assumption \eqref{eq:assumptionlemma2} implies that $(u_{k})$ is a decreasing sequence. If the sequence reaches $0$, the claim becomes trivial, so we would exclude this case in the proof, and focus on $k$ such that $u_{k}>0$. Exploiting the assumption in \eqref{eq:assumptionlemma2}: 
\begin{equation}\label{eq:lemma2e1}
\frac{1}{u_{k}}-\frac{1}{u_{k-1}}=\frac{u_{k-1}-u_{k}}{u_{k-1}u_{k}}\geq \mu \frac{u_{k-1}}{u_{k}} \geq \mu.
\end{equation}
By summing these inequalities, the following inequality holds:
$$\forall k \leq k_{0}, \quad \frac{1}{u_{k}}\geq \frac{1}{u_{0}} +k\mu \geq \frac{1+k\mu u_{0} }{u_{0}}  $$
It entails the claim $(i)$ of the Lemma.
    \item[$(ii)$] Transposing \eqref{eq:lemma2e1} to the probabilistic scenario, the assumption $\eqref{eq:assumptionlemma2proba}$ provides the following inequality:
   \begin{equation*}
   \prob \left( \frac{1}{u_{k}}-\frac{1}{u_{k-1}} \geq \mu  \right) > 1- \iota
    \end{equation*}
By considering the intersection of these events, we deduce:
   \begin{equation*}
    \prob \left( \frac{1}{u_{k}}-\frac{1}{u_{0}} \geq k \mu \right) \geq  \prob \left( \cap_{l=0}^{k} \left(\frac{1}{u_{l}}-\frac{1}{u_{l-1}} \geq  \mu \right) \right) \geq (1- \iota)^{k}
    \end{equation*}
\end{enumerate}
We can easily conclude.
\end{proof}

Let $\beta^{\musFlat{}}$ denote a lower bound for the average length of an episode, and $\alpha^{\musFlat{}}$ (resp. $ \alpha^{\musSharp{}}$) a lower (resp. upper bound) on the learning rate.
\begin{theorem}\label{cvtheorem}
\begin{enumerate}
    \item[$(i)$] Under Assumptions (\ref{as:hypswf}, \ref{as:hypAbound}, \ref{as:hypJ}, \ref{as:hypR}, \ref{as:hypalpha}, \ref{as:hyppi}, \ref{as:hypJhat}, \ref{as:hypAhat}, \ref{as:hypAinter}), and assuming the size of the minibatch $m$ tends towards infinity, the social welfare objective $\mathfrak{J}(\bm \theta^{k})$ converges in probability and with a sublinear convergence rate \paul{within} a radius of convergence $\tilde{\mathfrak{r}}$ \paul{of} the optimal value \paul{where}:
\begin{equation*}
\tilde{\mathfrak{r}}=\frac{ \Xi \left( 2 R_{0} +\alpha^{\musSharp{}}\right)}{2\beta^{\musFlat{}}} , \quad \text{ with } \quad  \Xi\coloneqq (L_{\swf} B_{A} \Xi_{J} + M_{\swf} \Xi_{A} + M_{\swf}  \Xi_{L}) M_{\bm \pi}.
\end{equation*}
More precisely, \paul{the} suboptimality after $k$ iterations may be bounded:
$$\lim_{m \rightarrow \infty} \prob \left( \left( \mathfrak{J}^{\ast}-\mathfrak{J}(\bm \theta^{k}) \right)  \leq\max \left( \tilde{\mathfrak{r}} ,  \frac{\mathfrak{j}_{0}}{\kappa \mathfrak{j}_{0} k+1}\right) \right) =1 \quad$$
$$\text{  with  } \mathfrak{j}_{0} \coloneqq \mathfrak{J}^{\ast}-\mathfrak{J}(\bm \theta^{0}) \text{ , } \kappa\coloneqq \frac{\upsilon-1}{\upsilon}\frac{\alpha^{\musFlat{}} \beta^{\musFlat{}}}{2 R_{0}^{2}} \text{ , and  }   \upsilon\coloneqq\left(1+\frac{\alpha^{\musSharp{}}}{2R_{0}}\right) ^{2}
$$

  \item[$(ii)$] Under Assumptions (\ref{as:hypswf}, \ref{as:hypAbound}, \ref{as:hypJ}, \ref{as:hypR}, \ref{as:hypalpha}, \ref{as:hyppi}, \ref{as:hypJhat}, \ref{as:hypAhat}, \ref{as:hypAinter}), the social welfare objective $\mathfrak{J}(\bm \theta^{k})$ converges, with a certain probability, \paul{within} the radius of convergence $\tilde{\mathfrak{r}}$ \paul{of} the optimal value. \\
  More precisely, \paul{the} suboptimality after $k$ iterations may be bounded:
\begin{equation}\label{eq:suboptimialitybound}
\forall \epsilon >0, \forall k>0, \quad
\prob \left( \mathfrak{J}^{\ast}-\mathfrak{J}(\bm \theta^{k}) \leq \max \left( \tilde{\mathfrak{r}}_{\epsilon},  \frac{\mathfrak{j}_{0}}{\kappa\mathfrak{j}_{0} k+1} \right) \right) \geq 1- k \iota_{\epsilon, m}  + o(\iota_{\epsilon, m}),
\end{equation}
\begin{equation*}
\text{  with  }  \tilde{\mathfrak{r}}_{\epsilon}\coloneqq\frac{ (\Xi+\epsilon) \left(  2R_{0} +\alpha^{\musSharp{}}\right)}{2\beta^{\musFlat{}}}.
\end{equation*}
 
 \item[$(iii)$] Similarly, under Assumptions (\ref{as:hypswf}, \ref{as:hypAbound}, 3', \ref{as:hypalpha}, \ref{as:hyppi}, \ref{as:hypJhat}, \ref{as:hypAhat}, \ref{as:hypAinter}), if $\vJ$ is simply locally concave, the SWO converges in probability and with a sublinear rate of convergence within the radius $\tilde{\mathfrak{r}}$ \paul{of} a local optimum, under the limit where $m$ tends toward infinity. Under these same assumptions, in the general case, we can, as in \eqref{eq:suboptimialitybound} bound the suboptimality relative to a local optima with a certain probability.

\end{enumerate}
\end{theorem}
\paul{This theorem} implies that we can bound the number of iterations required to have $\norm{\mathfrak{J}(\bm \theta^{k})} < \epsilon$ in probability:
\begin{corollary}\label{cvcoro}
Under Assumptions (\ref{as:hypswf}, \ref{as:hypAbound}, \ref{as:hypJ}, \ref{as:hypR}, \ref{as:hypalpha}, \ref{as:hyppi}, \ref{as:hypJhat}, \ref{as:hypAhat}, \ref{as:hypAinter}), with the previous notations:
\begin{equation*}
\forall \epsilon \text{ s.t } \tilde{\mathfrak{r}}<\epsilon<\mathfrak{j}_{0} , \quad \forall k \geq k_{\epsilon}, \quad \lim_{m \rightarrow \infty}  \prob \left( \mathfrak{J}^{\ast}-\mathfrak{J}(\bm \theta^{k}) < \epsilon \right)=1 , \quad \text{ with } k_{\epsilon}\coloneqq\frac{\upsilon}{\upsilon-1}\frac{2R_{0}^{2}}{\alpha^{\musFlat{}}\beta^{\musFlat{}}} \left( \frac{1}{\epsilon} - \frac{1}{\mathfrak{j}_{0}}\right).
\end{equation*}
\end{corollary}
Assuming that the terms $\Xi_{\vJ}, \Xi_{A}, \Xi_{L}$ stemming from Assumptions \ref{as:hypJhat}, \ref{as:hypAhat}, \ref{as:hypAinter} may be as small as desired, the convergence radius would tend towards 0, under the limit $m$ tends to infinity. 
However, in case of interdependent agents, particularly in cooperative or competitive settings, it is not reasonable to neglect the term stemming from Assumption $\ref{as:hypAinter}$. 
In contexts where \paul{the} interdependence between agents is tamer, these different advantages (of an agent depending on other agents actions) could average themselves out in \eqref{eq:AjiSWF}.

\begin{proof}[Proof of Corollary \ref{cvcoro}] 
Directly following from the assertion $(i)$ of Theorem \ref{cvtheorem}, since it is straightforward to check that $\forall k\geq k_{\epsilon}$, 
$\frac{\mathfrak{j}_{0}}{\kappa \mathfrak{j}_{0} k +1} \leq \epsilon$.
\end{proof}

\begin{proof}[Proof of Theorem \ref{cvtheorem}]  
As previously stated in Lemma \ref{jswfconcave}, under Assumptions (\ref{as:hypswf}, \ref{as:hypJ}), $\mathfrak{J}$ is concave.
From the concavity of $\mathfrak{J}$, applying Cauchy$-$Schwartz inequality: 
  \begin{equation}\label{eq:6}
   \mathfrak{J}^{\ast}-  \mathfrak{J}(\bm \theta^{k}) \leq \norm{ \nabla \mathfrak{J}(\bm \theta^{k}) } \norm{\bm \theta^{\ast}- \bm \theta^{k} }\leq R_{0}  \norm{\nabla \mathfrak{J}(\bm \theta^{k}) },
\end{equation}
where the last inequality stems from Assumption \ref{as:hypR}. 
Indeed, Assumption \ref{as:hypR} is satisfied for $\bm \theta^{0}$ and it is easy to prove by recursion, that it holds for all \paul{the} following $\bm \theta^{k}$'s; given \eqref{eq:JK}, we can prove recursively $\mathfrak{J}(\bm \theta^{k}) \geq \mathfrak{J}(\bm \theta^{0})$.

Combining \eqref{eq:6} with Lemma \ref{lemma1}:
\begin{equation}\label{eq:JK}
     \mathfrak{J}(\bm \theta^{k+1})- \mathfrak{J}(\bm \theta^{k}) 
     \geq \kappa_{k}  ( \mathfrak{J}(\bm \theta^{k})-  \mathfrak{J}^{\ast})^{2}  -   \mathcal{E}_{k}= \kappa_{k}  \Big[ ( \mathfrak{J}(\bm \theta^{k})-  \mathfrak{J}^{\ast})^{2} - \mathfrak{d}_{k}^{2}\Big]  \quad \text{ where } \left\{ \begin{array}{lll}
         \kappa_{k} &= & \frac{\alpha_{k} \beta_{k}}{2 R_{0}^{2}}  \\
         \mathcal{E}_{k} &=& \frac{\alpha_{k}}{2\beta_{k}} \norm{\err^{k}}^{2}\\
         \mathfrak{d}_{k}&=& \sqrt{\frac{\mathcal{E}_{k}}{\kappa_{k}}}  =\frac{\norm{\err^{k}} R_{0}}{\beta_{k}}
     \end{array} \right. .
\end{equation}

Rewriting \eqref{eq:JK}, with  $\mathfrak{j}_{k}=\mathfrak{J}^{\ast}-\mathfrak{J}(\bm \theta^{k})$:
\begin{equation}\label{eq:jK}
\mathfrak{j}_{k}-\mathfrak{j}_{k+1} \geq \kappa_{k} ( \mathfrak{j}_{k}^{2} - \mathfrak{d}_{k}^{2}),
\end{equation}

Below, we first proceed to the convergence analysis while assuming the error term $\norm{\err^{k}}$ is bounded by $\tilde{\Xi}$, in order to familiarise the reader with the proof's argument. In a second step, we refine this analysis, in order to work in probability. 
\paragraph{Prior Analysis:}
In this prior analysis, we assume the error term is bounded by $\tilde{\Xi}$. It enables us to define an upper bound for $\mathfrak{d}_{k}$:
\begin{equation}\label{eq:rdef}
 \mathfrak{d}_{k} \leq \mathfrak{r} \coloneqq \frac{\tilde{\Xi} R_{0}}{\beta^{\musFlat{}}} 
\end{equation}
From \eqref{eq:JK}, we deduce that the sequence $ (\mathfrak{J}(\bm \theta^{k+1}))_{k}$ is increasing at least until it reaches a distance $\mathfrak{r}$ from the optimal value $\mathfrak{J}^{\ast}$.
Moreover, by setting $\mathfrak{r}^{\upsilon}\coloneqq\sqrt{\upsilon} \mathfrak{r}$, combining \eqref{eq:jK} and \eqref{eq:rdef}, we deduce:
\begin{equation}\label{eq:jk}
\forall \upsilon>1,  \forall k, \quad  \mathfrak{j}_{k} \geq \mathfrak{r}^{\upsilon} \Rightarrow
\mathfrak{j}_{k}-\mathfrak{j}_{k+1} \geq  \kappa_{k} \left( \mathfrak{j}_{k}^{2} -\frac{(\mathfrak{r}^{\upsilon})^{2}}{\upsilon}\right) \geq \frac{\upsilon-1}{\upsilon} \kappa_{k} \mathfrak{j}_{k}^{2} \geq \kappa^{\upsilon} \mathfrak{j}_{k}^{2}, \quad  \text{ with  } \kappa^{\upsilon} \coloneqq \frac{\upsilon-1}{\upsilon}  \frac{\alpha^{\musFlat{}} \beta^{\musFlat{}}}{2 R_{0}^{2}}.
\end{equation}

From there, applying Lemma \ref{lemma2} would enable us to roughly conclude that $\mathfrak{j}_{k}$ converges with a sublinear rate of convergence towards the optimal value, at least until it reaches $\mathfrak{r}$.
More precisely, fixing $k^{\upsilon}$ such that  $\mathfrak{j}_{k}> \mathfrak{r}^{\upsilon}$ for all $k<k^{\upsilon}$, Lemma \ref{lemma2} asserts:
\begin{equation}\label{eq:jkrate}
\forall k\leq k^{\upsilon}, \quad
\mathfrak{j}_{k} \leq \frac{\mathfrak{j}_{0}}{\kappa^{\upsilon}\mathfrak{j}_{0} k+1}
\end{equation}

Since this is valid for all $\upsilon>1$, and $\lim_{\upsilon \rightarrow 1} \mathfrak{r}^{\upsilon}= \mathfrak{r}$, it suggests the sequence $\mathfrak{j}_{k}$ converges towards the radius $\mathfrak{r}$. However, this previous radius of convergence $\mathfrak{r}$ is not guaranteed to be stable. Indeed, once reached the radius $\mathfrak{r}$, the right side of \eqref{eq:jK} would be negative, and we can not conclude the sequence $\left(\mathfrak{j}_{k}\right)$ is still decreasing.
However, we can rethink a stable radius of convergence.
Let us choose $k_{1}$ \paul{---}assuming it exists\paul{---} such that $\mathfrak{j}_{k_{1}}$ lies within the radius $\mathfrak{r}$, but $\mathfrak{j}_{k_{1}-1}$ does not. 
Then, rewriting \eqref{eq:jK}:
\begin{equation}\label{eq:jKtwist}
    \mathfrak{j}_{k_{1}}- \mathfrak{j}_{k_{1}-1}
     \leq \kappa_{k_{1}-1}  \left(  \mathfrak{d}_{k_{1}-1}^{2}- \mathfrak{j}_{k_{1}-1}^{2}  \right)
\end{equation}
Therefore, from the previous definitions and assumptions:
\begin{equation}\label{eq:7}
\norm{\mathfrak{j}_{k_{1}}} \leq\norm{ \mathfrak{j}_{k_{1}-1} } +
     \kappa_{k_{1}-1}  \norm{  \mathfrak{d}_{k_{1}-1}^{2}- \mathfrak{j}_{k_{1}-1}^{2} } \leq \mathfrak{r} +  \kappa_{k_{1}-1} \mathfrak{d}_{k_{1}-1}^{2} = \mathfrak{r} + \mathcal{E}_{k_{1}-1} \leq \frac{\tilde{\Xi}}{\beta^{\musFlat{}}} R_{0} + \frac{\alpha_{k_{1}-1}}{2\beta_{k_{1}-1}} \tilde{\Xi}\leq \tilde{\mathfrak{r}},
\end{equation}
where $\alpha^{\musSharp{}}$ is an upper bound over the learning rate, and with our redefined radius of convergence:
$$\tilde{\mathfrak{r}}\coloneqq \frac{\tilde{\Xi}}{\beta^{\musFlat{}}}\left( R_{0}+\frac{\alpha^{\musSharp{}}}{2} \right).$$
By \eqref{eq:7}, $\mathfrak{j}_{k_{1}}$ lies within this radius. 
We claim that
all the following values $(\mathfrak{j}_{k})_{k>k_{1}}$ stay within the radius $\tilde{\mathfrak{r}}$. 
Indeed, by \eqref{eq:jK}, we can deduce that if $\mathfrak{j}_{k_{1}}>\mathfrak{r}$, the successive values $\left(\mathfrak{j}_{k}\right)_{k>k_{1}}$ will be again decreasing at least until reaching the radius $\mathfrak{r}$. 
Therefore, even when some values in the sequence $\left(\mathfrak{j}_{k}\right)$ ventures out of the original radius of convergence $\mathfrak{r}$, they will stay within the new radius of convergence $\tilde{\mathfrak{r}}$.
\paragraph{Analysis in Probability} In order to finalise the theorem proof, we will have to adapt the above arguments in probability. First, let us examine the error term, which, from the decomposition \eqref{eq:err}, satisfies:
\begin{equation*}
\norm{\err^{k}} \leq \norm{\bm{\nu}^{k}} +  \norm{\bm{\eta}^{k}}
\end{equation*}
A consequence of \paul{the} Chebyshev inequality, upon assumption \ref{as:hyppi} bounding the variance of the estimated gradient:
\begin{equation*}
\forall \epsilon>0, \quad  \prob \left( \norm{\bm{\nu}^{k}} <\epsilon\right)=  \prob  \left(\norm{ \hat{\gr}^{k}- \tilde{\gr}^{k}}  <\epsilon\right)\geq 1-\frac{ (\sigma^{\musSharp{}})^{2}}{\epsilon^{2}m},
\end{equation*}
where $m$ is the size of the minibatch on which the gradient $\hat{\gr}^{k}$ is estimated.
Meanwhile, as demonstrated in Lemma \ref{lemmaeta}, $\norm{\bm{\eta}^{k}}$ may be bounded by $\Xi$ in probability, under Assumptions \ref{as:hypswf}, \ref{as:hypAbound}, \ref{as:hyppi}, \ref{as:hypJhat}, \ref{as:hypAhat}, \ref{as:hypAinter}.
Hence:
\begin{equation}\label{eq:errbound1}
\forall \epsilon>0, \quad  \prob \left( \norm{\err^{k}} <\Xi_{\epsilon}\right)\geq 1-\iota_{\epsilon, m}, \quad \text{ with } \iota_{\epsilon, m}\coloneqq \frac{ (\sigma^{\musSharp{}})^{2}}{\epsilon^{2}m}, \text{ and }  \Xi_{\epsilon}\coloneqq \Xi + \epsilon
\end{equation}
In particular, upon taking the limit over $m$:
\begin{equation}\label{eq:errbound}
 \forall \epsilon>0, \quad \lim_{m\rightarrow\infty} \prob \left( \norm{\err^{k}} <\Xi_{\epsilon}\right)=1.
\end{equation}
Instead of the $\eqref{eq:rdef}$ from the prior analysis, we get the following probabilistic form:
\begin{equation}\label{eq:rdefproba}
\forall \epsilon>0, \quad \prob\left( \mathfrak{d}_{k} \leq \mathfrak{r}_{\epsilon} \right) \geq 1-\iota_{\epsilon, m}, \quad \text{ where } \mathfrak{r}_{\epsilon} \coloneqq \frac{\Xi_{\epsilon} R_{0}}{\beta^{\musFlat{}}}.
\end{equation}
By setting $\mathfrak{r}_{\epsilon}^{\upsilon}\coloneqq\sqrt{\upsilon} \mathfrak{r}_{\epsilon}$, it induces the following ---since $\mathfrak{j}_{k}^{2} \geq \upsilon \mathfrak{d}_{k}^{2} \Rightarrow  \mathfrak{j}_{k}^{2}-\mathfrak{d}_{k}^{2} \geq \mathfrak{j}_{k}^{2}\frac{\upsilon-1}{\upsilon} $:
\begin{equation*}
\forall \epsilon>0, \forall \upsilon>1, \forall k, \quad \prob \left(
\mathfrak{j}_{k}^{2} -\mathfrak{d}_{k}^{2} \geq \frac{\upsilon-1}{\upsilon} \mathfrak{j}_{k}^{2} \mid  \mathfrak{j}_{k} \geq \mathfrak{r}_{\epsilon}^{\upsilon}  \right) \geq
\prob \left(
\mathfrak{j}_{k}^{2} \geq \upsilon \mathfrak{d}_{k}^{2}\mid \mathfrak{j}_{k} \geq \mathfrak{r}_{\epsilon}^{\upsilon}  \right) \geq \prob \left( \mathfrak{d}_{k} \leq \mathfrak{r}_{\epsilon}\right) \geq 1-\iota_{\epsilon, m}.
\end{equation*}

With such an inequality in hand, we deduce from \eqref{eq:jK}:
\begin{equation}\label{eq:jkproba}
\forall \epsilon>0, \forall \upsilon>1, \quad \prob \left(
\mathfrak{j}_{k}-\mathfrak{j}_{k+1} \geq \kappa^{\upsilon} \mathfrak{j}_{k}^{2} \mid \mathfrak{j}_{k} \geq \mathfrak{r}_{\epsilon}^{\upsilon} \right) \geq 1-\iota_{\epsilon, m}, \text{ where } \kappa^{\upsilon}\coloneqq\frac{\upsilon-1}{\upsilon} \frac{\alpha^{\musFlat{}} \beta^{\musFlat{}}}{2 R_{0}^{2}}
\end{equation}
Notably, passing to the limit over $\upsilon$ in \eqref{eq:jkproba}, since $\mathfrak{r}_{\epsilon} =\lim_{ \upsilon \rightarrow 1} \mathfrak{r}_{\epsilon}^{\upsilon}$:
\begin{equation}\label{eq:decreaseproba}
\prob \left(
\mathfrak{j}_{k} \geq \mathfrak{j}_{k+1} \mid \mathfrak{j}_{k} \geq \mathfrak{r}_{\epsilon} \right)\geq 1- \iota_{\epsilon, m}
\end{equation}
The sequence $\mathfrak{j}_{k}$ is therefore decreasing until it reaches $\mathfrak{r}_{\epsilon}$, with a certain probability. At the limit where $m\rightarrow \infty$, $\mathfrak{j}_{k}$ is almost surely decreasing, until reaching the radius $\mathfrak{r}$, since we can tie tying $\epsilon$ to $m$ such that $\lim_{m\rightarrow \infty, \epsilon\Rightarrow \infty} \iota_{epsilon, m}=0$, e.g. with $\epsilon=m^{-\frac{1}{3}}$.
Applying Lemma \ref{lemma2} to the last inequality would enable us to roughly conclude that $\mathfrak{j}_{k}$ converges with a sublinear rate a convergence with a certain (bounded) probability until it reaches a radius $\mathfrak{r}_{\epsilon}$.

More precisely, let us fix $k^{\upsilon}_{\epsilon}$ such that $\forall k< k^{\upsilon}_{\epsilon}, \mathfrak{j}_{k}> \mathfrak{r}_{\epsilon}^{\upsilon}$. Then, similarly to $\eqref{eq:jkrate}$, Lemma \ref{lemma2} leads to:
\begin{equation}\label{eq:jkrateproba}
\forall k\leq k^{\upsilon}_{\epsilon}, \quad
\prob \left( \mathfrak{j}_{k} \leq  \frac{\mathfrak{j}_{0}}{\mathfrak{j}_{0} \kappa^{\upsilon} k+1} \right) \geq \left(1-\iota_{\epsilon, m}\right)^{k}= 1 -k \iota_{\epsilon, m} + o(\iota_{\epsilon, m})
\end{equation}
As this is valid for all $\upsilon>1$ (although it changes the factor $\kappa^{\upsilon}$ of the convergence rate), it signifies the sequence $\mathfrak{j}_{k}$ is decreasing until $k$ and converges towards $\mathfrak{r}_{\epsilon}$ with a certain probability.

However, as highlighted previously in $(i)$, this radius $\mathfrak{r}_{\epsilon}$ is not guaranteed to be stable. Let us fix $k_{0, \epsilon}$ the first $k$ such that $\mathfrak{j}_{k}\leq \mathfrak{r}_{\epsilon}$ and $k_{1, \epsilon}$ the first $k>k_{0, \epsilon}$, such that $\mathfrak{j}_{k}> \mathfrak{r}_{\epsilon}$ again ---assuming they exists. Similarly to \eqref{eq:7}, from \eqref{eq:jKtwist}:
\begin{equation}\label{eq:7proba}
\mathfrak{j}_{k_{1, \epsilon}} \leq \mathfrak{j}_{k_{1, \epsilon}-1}  +
     \kappa_{k_{1, \epsilon}-1}  \norm{  \mathfrak{d}_{k_{1, \epsilon}-1}^{2}- \mathfrak{j}_{k_{1, \epsilon}-1}^{2} } \leq \mathfrak{r}_{\epsilon} +  \kappa_{k_{1, \epsilon}-1} \mathfrak{d}_{k_{1, \epsilon}-1}^{2}= \mathfrak{r}_{\epsilon} + \mathcal{E}_{k_{1, \epsilon}-1},
\end{equation}
From the assumption on the error term and definition of $\mathcal{E}_{k}$, with the previous notations:
\begin{equation}\label{eq:8proba}
\forall k, \quad \prob \left(\mathcal{E}_{k} < \frac{\alpha^{\musSharp{}}}{2 \beta^{\musFlat{}}}\Xi_{\epsilon} \right) > 1 - \iota_{\epsilon, m}
\end{equation}
Combining \eqref{eq:7proba} and \eqref{eq:8proba}, and introducing the new convergence radius, it implies:
\begin{equation*}
\prob \left(\mathfrak{j}_{k_{1, \epsilon}} \leq  \tilde{\mathfrak{r}}_{\epsilon} \right) \geq 1 - \iota_{\epsilon, m}, \text{ where } \tilde{\mathfrak{r}}_{\epsilon}\coloneqq \frac{\Xi_{\epsilon}}{\beta^{\musFlat{}}}\left(R_{0}+\frac{\alpha^{\musSharp{}}}{2} \right).
\end{equation*}
By \eqref{eq:jkproba} and the initial assumption on $k_{1, \epsilon}$, the sequence $\mathfrak{j}_{k}$ is now decreasing with probability $(1 - \iota_{\epsilon, m})$.
Thereupon, the following value $\mathfrak{j}_{k_{1, \epsilon}+1}$ satisfies:
\begin{equation*}
\prob \left(\mathfrak{j}_{k_{1, \epsilon}+1} \leq  \tilde{\mathfrak{r}}_{\epsilon} \right) \geq \prob \left(\mathfrak{j}_{k_{1, \epsilon}} \leq  \tilde{\mathfrak{r}}_{\epsilon} \right)  \prob \left(\mathfrak{j}_{k_{1, \epsilon}+1} \leq  \mathfrak{j}_{k_{1, \epsilon}} \right) \geq (1 - \iota_{\epsilon, m})^{2} 
\end{equation*}
Iterating such argument, would lead us to the following bound:
\begin{equation}\label{eq:jkrate2proba}
\forall k > k_{1, \epsilon}, \quad 
\prob \left(\mathfrak{j}_{k} \leq  \tilde{\mathfrak{r}}_{\epsilon} \right) \geq (1 - \iota_{\epsilon, m})^{k-k_{1, \epsilon}} \geq (1 - \iota_{\epsilon, m})^{k}.
\end{equation}
This argument may be repeated for successive values of the sequence $\mathfrak{j}_{k}$, if such behaviors happens. 
Gathering the previous results,  \eqref{eq:jkrateproba} and \eqref{eq:jkrate2proba}, leads us to the following claim:
\begin{equation}\label{eq:jkrateprobafinal}
\forall \epsilon >0, \forall k>0, \quad
\prob \Big[ \mathfrak{j}_{k} \leq \max \left( \tilde{\mathfrak{r}}_{\epsilon},  \frac{\mathfrak{j}_{0}}{\kappa^{\upsilon}\mathfrak{j}_{0}k+1} \right) \Big]\geq \left(1-\iota_{\epsilon, m}\right)^{k}= 1 -k \iota_{\epsilon, m} + o(\iota_{\epsilon, m}),
\end{equation}
under the condition of $\upsilon$ satisfying: $$\mathfrak{r}_{\epsilon}^{\upsilon}\leq \tilde{\mathfrak{r}_{\epsilon}} \iff \sqrt{\upsilon}R_{0} \leq \left(R_{0}+\frac{\alpha^{\musFlat{}}}{2}\right) \iff \upsilon \leq \left(1+ \frac{\alpha^{\musFlat{}}}{2R_{0}}\right)^{2}. $$ For the theorem, we adopted $\upsilon=\left(1+ \frac{\alpha^{\musFlat{}}}{2R_{0}}\right)^{2}$.

\paragraph{Limit case:}
As previously mentionned, by taking the limit $m \rightarrow \infty$, and $\epsilon \rightarrow 0$, such that $\epsilon^{2}m \rightarrow \infty$ (e.g. $\epsilon = m^{-\frac{1}{3}}$), $\iota_{\epsilon, m}$ would tend towards $0$. Hence, \eqref{eq:jkrateprobafinal} reveals that $\mathfrak{j}_{k}$ converges in probability within the radius $\tilde{\mathfrak{r}}$ with a sublinear rate of convergence, as claimed.

\paragraph{Locally Concave Case:}
The case of assuming $\vJ$ simply locally concave would lead to a similar convergence result, under the corresponding set of assumptions, yet for a local optimum on the neighborhood $\mathcal{U}$. Since the proof is almost identical to the previous case, it is left to the reader.

\end{proof}

\section{Additional Experimental Results}
\label{app:moreexp}

\mz{
\us{We present here the additional experimental results that were not included in the main paper. Through those experiments, we try to answer the following questions:}

\newcounter{addquestions}
\setcounter{addquestions}{1}

\begin{itemize}
    \item {\bf (\Alph{addquestions})} How \pw{do} SOTO and FEN perform with unequal access to resources in a controlled domain?
    \stepcounter{addquestions}
    \item {\bf (\Alph{addquestions})} Can SOTO cope \pw{with} a different number of users and agents?
    \stepcounter{addquestions}
    \item {\bf (\Alph{addquestions})} What if Basic could also learn from $\hat{\bm A}^\text{IND}_i$ then $\hat{\bm A}^\text{SWF}$ gradually with $\beta$?
    \stepcounter{addquestions}
    \item {\bf (\Alph{addquestions})} What if we use the same architecture proposed in FEN to optimize a SWF?
    \stepcounter{addquestions}
    \item {\bf (\Alph{addquestions})} Where does the bad performance of FEN in some domains come from?
    \stepcounter{addquestions}
    \item {\bf (\Alph{addquestions})} Does SOTO still outperform FEN if it is also allowed to share scores to neighbors?
    \stepcounter{addquestions}
    \item {\bf (\Alph{addquestions})}
    \us{Is Basic capable enough to find fair solutions in small problems?}
\end{itemize}
\setcounter{addquestions}{0}

}

\paragraph{Question (\Alph{addquestions})}\stepcounter{addquestions}
In those experiments, we \pw{evaluate} how our method performs when there is an unequal access to resources.
We modified the Matthew \us{Effect} environment to restrict one of the 3 resources to be only collectible by the first two agents.
They have an additional observation to determine whenever the resource is restricted.
Other agents neither observe nor collect those restricted resources.

\begin{figure}[H]
    \centering
    \includegraphics[width=0.6\linewidth]{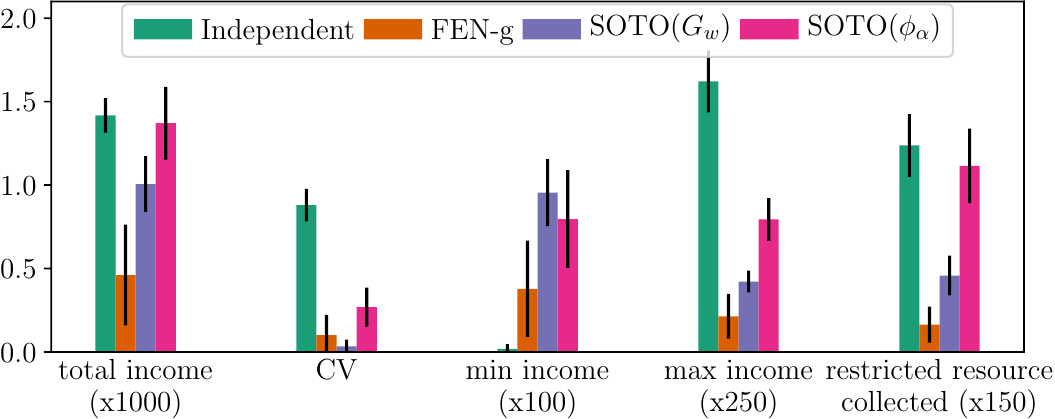}
    \caption{Comparison of Independent, FEN, and SOTO on Matthew Effect with unequal access to resources in the CLDE scenario.
    Higher total income is better, lower CV is better, higher min income is better, higher max income is better and higher restricted resource collected is better.}
    \label{fig:results_matthew_bar_restricted}
\end{figure}

\begin{figure}[H]
    \centering
    \includegraphics[width=\linewidth]{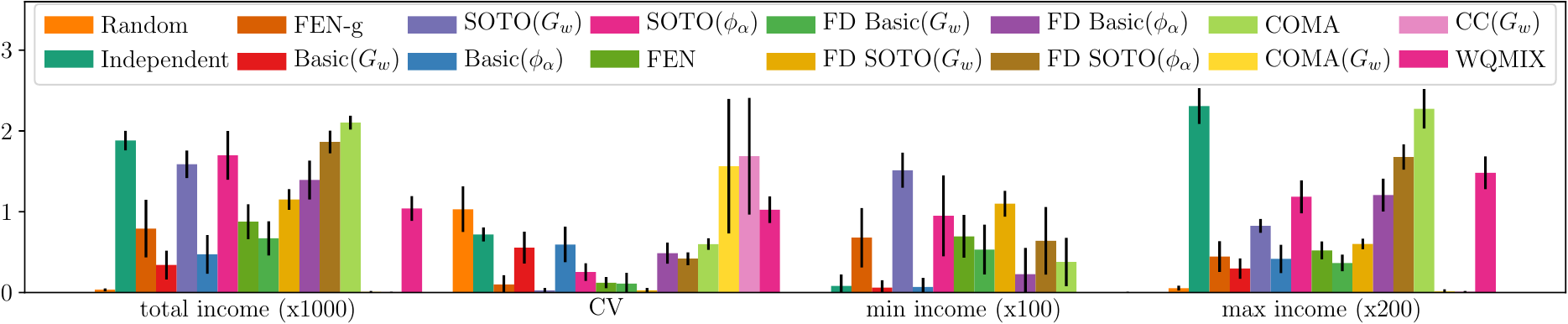}
    \caption{Comparison of different methods on Matthew Effect.}
    \label{fig:results_matthew_bar}
\end{figure}

\mz{Compared to Matthew Effect without resource restriction (Figure~\ref{fig:results_matthew_bar}), in Figure~\ref{fig:results_matthew_bar_restricted}, as expected,  Independent achieves a lower total income since it is more difficult to collect resources for the majority of the agents.
}
FEN is Pareto-dominated by SOTO($G_{\bm w}$). 
Because of the unequal access to resources, the CV criterion has less importance in this experience: the first two agents should collect the maximum of restricted resources.
This behavior is achieved by SOTO($\phi_\alpha$), which collected a high number of restricted resources, which also Pareto-dominates FEN if we ignore the CV criterion.
This shows that our method can cope with unequal access to resources \mz{whereas the first two agents of FEN do not collect more resources than the others}.

\paragraph{Question (\Alph{addquestions})}\stepcounter{addquestions}
\mz{
\us{To answer (\textbf{\Alph{addquestions}}), we turn to a more complex} SUMO environment where we split the objectives per road instead of per intersection. The objective is to minimize the waiting time of the two roads of an intersection, hence each agent observes two users ($2N=D$).
Accordingly, the output size of the critics in the SOTO architecture is two.
}

\begin{figure}[H]
    \centering
    \includegraphics[width=0.7\linewidth]{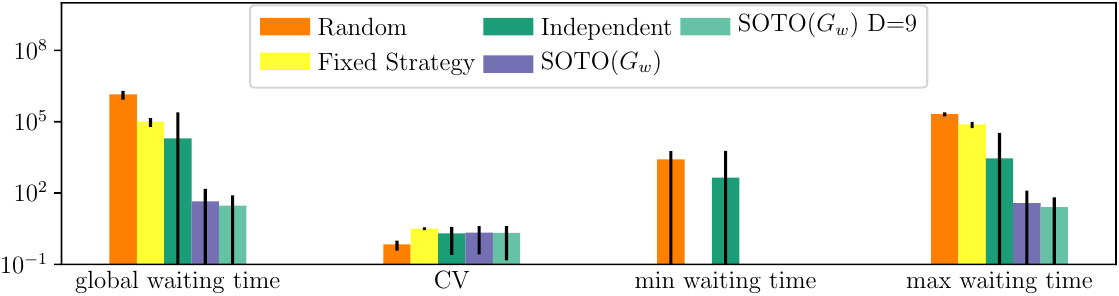}
    \caption{SUMO simulation with D=18 in the CLDE scenario. Lower global waiting time is better, lower CV is better, lower min waiting time is better and lower max waiting time is better.}
    \label{fig:sumo18D}
\end{figure}

In Figure \ref{fig:sumo18D}, we can see that our method copes \pw{with} a different number of users and agents.
\mz{However, this artificial splitting \pw{does not} allow for a fairer solution than the one when $D=9$.}

\paragraph{Question (\Alph{addquestions})}\stepcounter{addquestions}
\mz{To verify that the SOTO architecture is \us{necessary} and the benefits do not come only from training with $\beta$, we performed another control experiment where Basic is also trained with the two different objectives: $\hat{\bm A}^\text{IND}_i$ and then $\hat{\bm A}^\text{SWF}$.} 

\begin{figure}[H]
    \centering
    \includegraphics[width=0.45\linewidth]{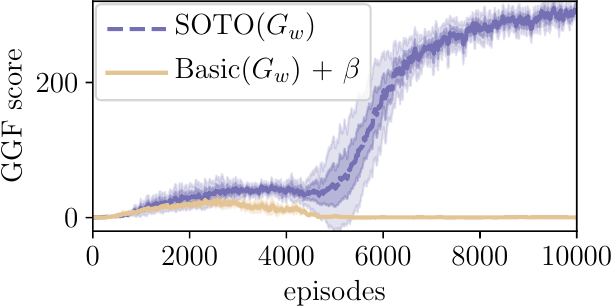}
    \caption{Comparison between SOTO and Basic updated by the two objectives with $\beta$ on Matthew Effect.}
    \label{fig:basicbeta}
\end{figure}

\mz{
Figure~\ref{fig:basicbeta} can be analyzed in two parts.
In the first stage, when $\hat{\bm A}^\text{IND}_i$ is mostly used to update the network, both approaches achieve an almost similar performance. 
However, the more $\hat{\bm A}^\text{SWF}$ is used, the more catastrophic forgetting intervenes.
This demonstrates the importance of having two separate networks for each of the objectives.
}

\paragraph{Question (\Alph{addquestions})}\stepcounter{addquestions}
\mz{Here we want to verify if we could use the FEN architecture instead of the SOTO one.
To do so, we \us{extend} FEN such that the meta-controller is rewarded by the current GGF value (FEN-g GGF).
We also extended FEN such that the meta-controller is updated with $\hat{\bm A}^\text{SWF}$ (FEN-g $A^\text{SWF}$).
}

\begin{figure}[H]
    \centering
    \includegraphics[width=0.6\linewidth]{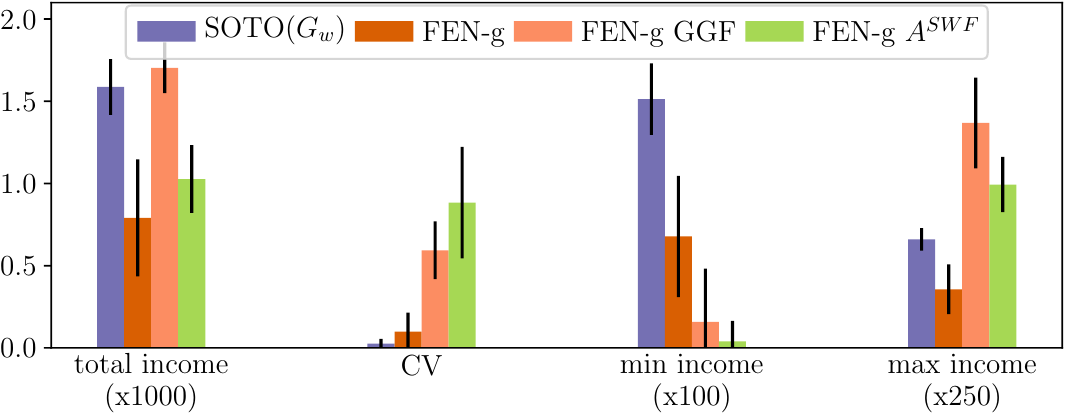}
    \caption{Comparison between SOTO, FEN without gossip, and FEN without gossip when it optimizes a SWF on the Matthew Effect. 
    Higher total income is better, lower CV is better, higher min income is better, higher max income is better and higher restricted resource collected is better.}
    \label{fig:fenggi}
\end{figure}

\mz{
Figure~\ref{fig:fenggi} \us{shows that} both baselines are able to learn a good behavior (in terms of total income).
\us{However, none of them learn to achieve the better CV than FEN-g and SOTO.}
From this experiment, we conclude again that the SOTO architecture is essential to optimize a SWF without a centralized critic.
}

\paragraph{Question (\Alph{addquestions})}\stepcounter{addquestions}

Since FEN was not able to learn a good policy on the SUMO domain, we also tried to perform a pretraining of the first sub-policy as suggested by the authors for Plant Manufacturing.
For the first 400 episodes, the first sub-policy is trained without interventions of the meta-controller.

\begin{figure}[H]
    \centering
    \includegraphics[width=0.6\linewidth]{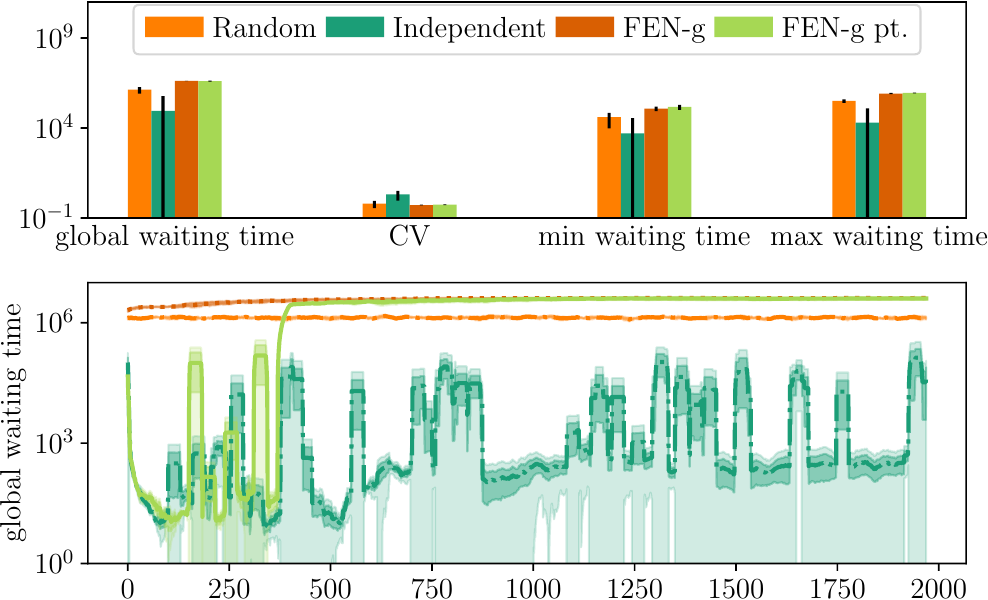}
    \caption{Comparison between FEN without gossip and FEN without gossip with pretraining (without gossip). Lower global waiting time is better, lower CV is better, lower min waiting time is better and lower max waiting time is better.}
    \label{fig:fendiverge}
\end{figure}

Figure~\ref{fig:fendiverge} shows that during the first 400 episodes, the performance of FEN and Independent are indeed similar.
However, when the meta-controller starts to switch other sub-policies the divergence begins.
Note that when the pretraining is over, the first sub-policy is not updated anymore.
So, if the performance drops it means that the meta-controller selects less and less often the first sub-policy.
This is probably due to the fact that several agents select a wrong sub-policy at the same time, so the average performance is greatly reduced. 
Therefore, the other agents who chose the first sub-policy will also choose bad sub-policies for the next time to reduce the CV leading to a vicious circle until the worst possible average is reached.
We observed this phenomenon in several domains.

\paragraph{Question (\Alph{addquestions})}\stepcounter{addquestions}
To verify that our better performances are not due to the fact that our networks have more input information, we provided FEN with the same additional inputs for its neural networks (i.e. the $\vJ(\bm \theta$) of the neighbors). 

\begin{figure}[H]
    \centering
    \includegraphics[height=4.cm]{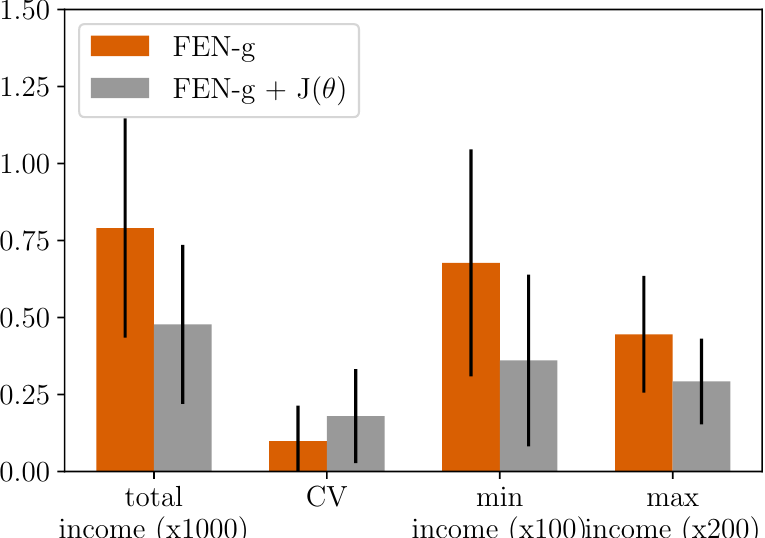}
    \includegraphics[height=4.cm]{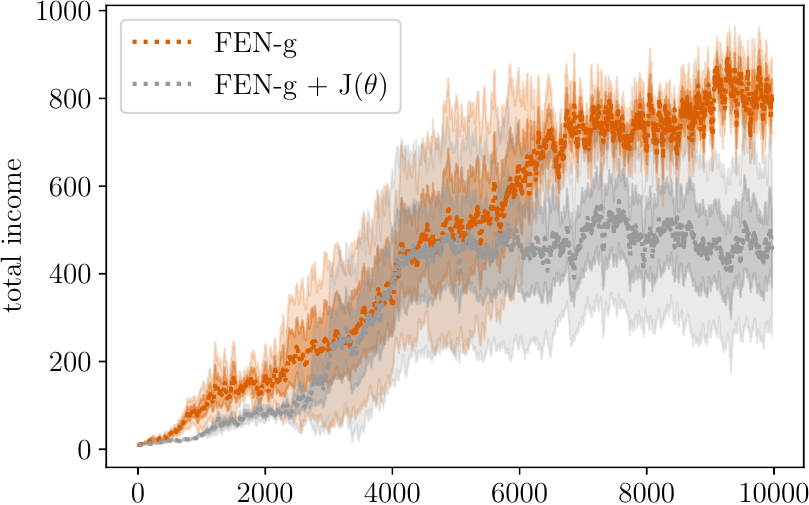}
    \caption{Comparison between FEN and FEN with more communications (without gossip).}
    \label{fig:fencom}
\end{figure}

From Figure~\ref{fig:fencom}, we can confirm that the additional communications are not helping FEN in Matthew Effect.

\us{
\paragraph{Question (\Alph{addquestions})}\stepcounter{addquestions}
In order to answer (\Alph{addquestions}), we discuss the experimental results of Independent, FEN, Basic and SOTO on Job Scheduling in the CLDE scenario.
Figure~\ref{fig:job_basic_vs_soto} illustrates the performances in terms of resource utilization, CV, minimum and maximum of utilities. 
As expected, Independent performs worse as it has the highest CV.
Our algorithms achieve lower CV than FEN which indicates Basic and SOTO methods are able to find more fairer solution than FEN.
Among all the algorithms, Basic($G_{\bm w}$) performs the best as it has the lowest CV.
From these results, we can conclude that in simple domains Basic can find optimal solution without our proposed neural network architecture while in difficult task such as in Matthew Effect~\ref{fig:results_matthew_bar} and Plant Manufacturing~\ref{fig:plant} it performs worse.

\begin{figure}[H]
    \centering
    \includegraphics[width=0.6\linewidth]{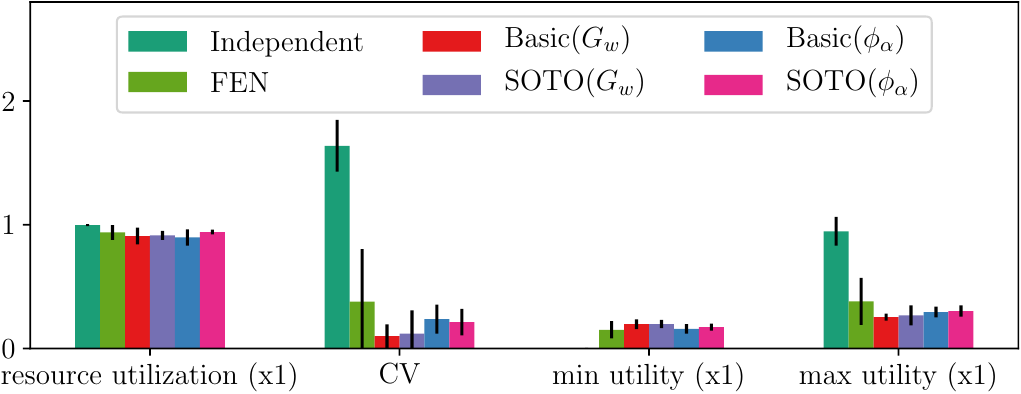}
    \caption{Comparison of Independent, FEN, Basic and SOTO on Job Scheduling in the CLDE scenario.}
    \label{fig:job_basic_vs_soto}
\end{figure}

}

\section{Complete Experimental Results Per Domain} \label{app:allexp}

In this part, we provide the results on the Job Scheduling, \us{Matthew Effect and Plant Manufacturing environments (Table~\ref{tab:results_job}, Figure~\ref{fig:results_matthew_bar} and Figure \ref{fig:plant})}.
\us{We also present an additional} plot to visualize the solutions on Matthew Effect in the FD scenario (Figure~\ref{fig:results_matthew_pareto_fd}).

\subsection{Job Scheduling}
\label{app:jobscheduling}

\begin{table}[H]
\begin{center}
\caption{Job scheduling benchmark. The first part refers to the CLDE scenario. The second part refers to the FD scenario. 
}
\label{tab:results_job}
\vskip 0.15in
\begin{tabular}{lcccccr}
\toprule
Method & Resource utilization & CV & min utility & max utility \\ 
\midrule
Independent & 1.00 $\pm$0.00 & 1.64 $\pm$0.19 & 0.00 $\pm$0.00 & 0.95 $\pm$0.11 \\
FEN without gossip &  0.93 $\pm$0.07 & 0.37 $\pm$0.42 & 0.15 $\pm$0.07 & 0.37 $\pm$0.19 \\
Basic($G_{\bm w}$) & 0.91 $\pm$0.07 & 0.10 $\pm$0.10 & 0.20 $\pm$0.04 & 0.25 $\pm$0.03 \\
SOTO($G_{\bm w}$) & 0.91 $\pm$0.04 &  0.12 $\pm$0.19 & 0.20 $\pm$0.03 & 0.27 $\pm$0.08 \\
Basic($\phi_\alpha$) & 0.90 $\pm$0.07 & 0.24 $\pm$0.12 & 0.16 $\pm$0.04 & 0.29 $\pm$0.04 \\
SOTO($\phi_\alpha$) &  0.94 $\pm$0.02 & 0.23 $\pm$0.10 & 0.16 $\pm$0.03 & 0.31 $\pm$0.04 \\ 
\midrule
Independent &  1.00 $\pm$0.00 & 1.64 $\pm$0.19 & 0.00 $\pm$0.00 & 0.95 $\pm$0.11 \\
FEN & 0.72 $\pm$0.24 & 0.84 $\pm$ 0.38 &  0.04 $\pm$0.03 & 0.44 $\pm$0.26 \\
FD Basic($G_{\bm w}$) & 0.79 $\pm$0.22 & 1.12 $\pm$0.46 & 0.02 $\pm$0.03 & 0.57 $\pm$0.28 \\
FD SOTO($G_{\bm w}$) & 0.87 $\pm$0.12 & 1.07 $\pm$0.42 & 0.02 $\pm$0.04 & 0.60 $\pm$0.23 \\
FD Basic($\phi_\alpha$) & 0.71 $\pm$0.31 & 1.36 $\pm$0.42 & 0.01 $\pm$0.02 & 0.61 $\pm$0.34 \\
FD SOTO($\phi_\alpha$) & 0.94 $\pm$0.11 & 1.61 $\pm$0.26 & 0.00 $\pm$0.00 & 0.89 $\pm$0.18 \\
\bottomrule
\end{tabular}
\end{center}
\vskip -0.1in
\end{table}


\subsection{Matthew Effect}

\subsection{Plant Manufacturing}
\begin{figure}[H]
    \centering
    \includegraphics[width=0.7\linewidth]{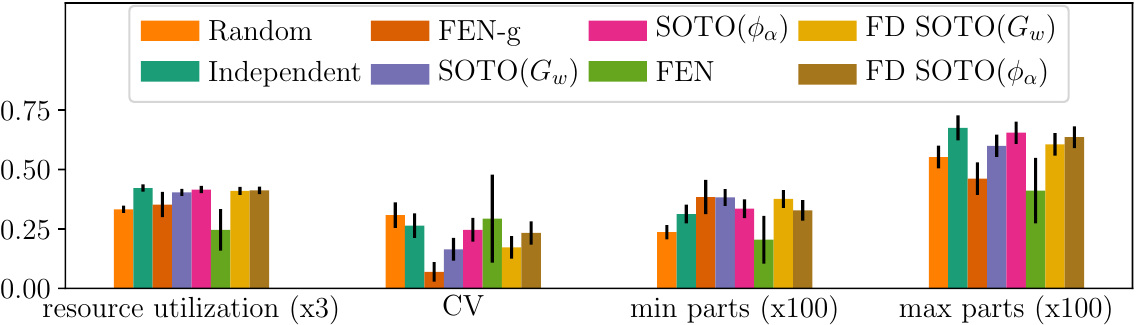}
    \caption{Comparison of different methods on Plant Manufacturing.}
   \label{fig:plant}
\end{figure}

In Plant Manufacturing, Figure~\ref{fig:plant}, in both scenarios, SOTO($G_{\bm w}$) and FEN are able to build the maximum number of products (equivalent to the minimum number of parts for each agent).

\section{More details on experimentation}
\label{app:envdetails}

Note that compared to the results presented in \citep{JiangLu19}, our Independent method works almost twice better because we normalize the inputs, used Generalized Advantage Estimation \citep{SchulmanMoritzLevineJordanAbbeel16} instead of pure Monte-Carlo, and added bias units to the neural networks. All the \us{presented} methods benefit from those improvements.

In both scenarios (CLDE and FD), FEN is not restricted to send messages to neighbors for its gossip algorithm.
To emphasize decentralization, all the agents are learning independent parameters without weights sharing (except for the COMA \us{and WQMIX} baseline\us{s}).

The source code is available at \url{https://gitlab.com/AAAL/DFRL} with the configuration files used to generate our experimental results.
The hyperparameters of the algorithms were optimized by grid search using Lightweight HyperParameter Optimizer (LHPO)\footnote{\url{https://github.com/matthieu637/lhpo}}, an open source library used to run asynchronous distributed experiments \cite{zimmer2018phd}.

\begin{figure}[H]
    \centering
    \includegraphics[width=0.55\linewidth]{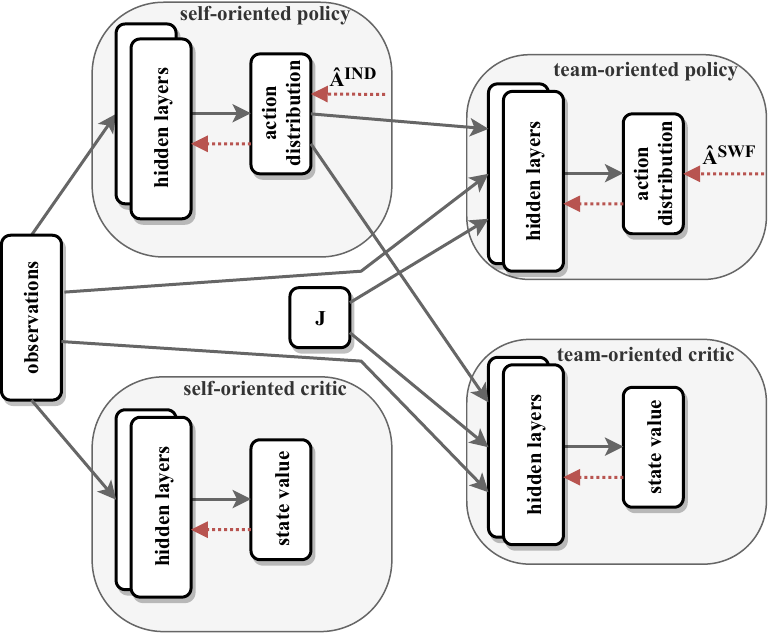}
    \caption{The complete SOTO architecture.}
    \label{fig:sotonnfull}
\end{figure}

The complete SOTO architecture (including critics) is provided in Figure~\ref{fig:sotonnfull}.

\subsection{Hyperparameters}  \label{app:hyper}
\us{For all the experiments, we use PPO.}
The hidden layers of the neural networks are composed of two ReLU layers with 256 units.
We used the ADAM \citep{kingma2014adam} optimization method with $2.5 \times 10^{-4}$ as learning rate for the actor and $1 \times 10^{-3}$ for the critic. 
The exploration bonus and the clipping ratio of the importance sampling in PPO are set to $0.03$ and $0.1$ respectively.
\mz{In both scenarios, the FEN gossip algorithm uses $g=10$ gossip rounds and $\tilde{k}=3$ agents receiving the message.}
In every environment, we used Generalized Advantage Estimation with $\lambda = 0.97$ except in the Job Scheduling environment where we kept pure Monte-Carlo estimation to reproduce the state-of-the-art results.
\us{The discount factor $\gamma$ is 0.98 for Plant Manufacturing, Matthew effect and Job Scheduling, 0.99 for SUMO and 0.95 for Iroko. 
Batch sizes are respectively 25, 50, 50, 50 and 128 in Job Scheduling, Matthew Effect, Plant Manufacturing, SUMO and Iroko.
In continuous action space (Iroko), we tuned the \mz{standard deviation of the Gaussian distribution} for each instance of algorithms \mz{instead of learning it (so it is independent of the state)}.
The best standard deviations for Independent, FEN-g, SOTO($G_{w}$) and FD SOTO$(G_{w})$ are 0.5, 0.5, 0.05, and 0.1 respectively.}

\subsection{Hardware}

We performed the experiments on different computers equipped with 2 x Intel Xeon CPU E5-2678 v3 or Intel Core i7-8700 for SUMO \mz{and Iroko} and 2 x Intel Xeon E5-2630 v3 or 2 x Intel Xeon Gold 6130 for the other domains.

\begin{table}[h]
\caption{Comparison of the average computation times in hours.}
\label{tab:cputime}
\vskip 0.15in
\begin{center}
\begin{tabular}{lccccccr}
\toprule
 Method & Job Scheduling & Matthew Effect & Plant Manufacturing & SUMO (D=9) & Iroko \\ \midrule
Independent & 9 & 27 & 8  & 20 & 5 \\
FEN-g & 13 & 37  & 64  & 60  & 6 \\
Basic($G_{\bm w}$) & 11 & 30  &  &   \\
SOTO($G_{\bm w}$) & 11 & 34 & 12 & 14 & 7 \\
Basic($\phi_\alpha$) & 11 & 23 &   &  \\
SOTO($\phi_\alpha$) & 14 & 29 & 11  & 27  \\ 
FEN & 14 & 34  &  80+  &  \\
FD Basic($G_{\bm w}$) & 11 & 26  &   &   \\
FD SOTO($G_{\bm w}$) & 13 &  27 & 12  & 14 & 7  \\
FD Basic($\phi_\alpha$) & 14  & 26  &  &   \\
FD SOTO($\phi_\alpha$) & 16 & 31  & 11  &   \\
Random &  & 4  & 8  & 45  & 4\\
Fixed Strategy &  &   &   & 45 & 4 \\
COMA & 13 & 23  &   &  &  \\
COMA($G_{\bm w}$) & 15 & 29  &   &  &  \\
CC($G_{\bm w}$) & 13 &  24 &   &  &  \\
WQMIX & 24 & 80+ &  & 70 & \\
\bottomrule
\end{tabular}
\end{center}
\vskip -0.1in
\end{table}

In Table~\ref{tab:cputime}, the large computation time of FEN in Plant Manufacturing is due to the fact that the episode can last 10000 steps if all the 800 gems have not been collected.

\subsection{Environments} \label{app:env}

\paragraph{Job Scheduling}
In Job Scheduling, a permanent and unique resource is placed on a grid with 4 agents, which they must learn to share it. 
In Job Scheduling, neighbors are limited to being one block away from the agent (the number of neighbors may vary over time).
The different neighborhoods correspond to the grid observations.

\paragraph{Plant Manufacturing}

In Plant Manufacturing, 3 types of gems are randomly placed in a grid along with 5 agents.
Once a gem is collected, it reappears at a random location with a random type.
Each agent needs a specific combination of the gem types to build a part.
Once each agent have a part, a product is built.
\mz{However, the agents are not aware of the complete definition of that reward function, they are rewarded for each gem and part, not for the full product.
}

In Plant Manufacturing, neighbors are limited to being two blocks away from the agent (the number of neighbors may vary over time).
The different neighborhoods correspond to the grid observations.

The reported parameter of the Plant Manufacturing between the original paper and the code provided by \citet{JiangLu19}  differed\footnote{Authors confirmed it in private communications. They also mentioned that a pretraining of the first sub-policy was required in this environment.}. Thus, we run all the algorithms with the following parameters: 800 gems are available for an episode, the reward bonus for collecting a gem is 0.01, the size of a minibatch is 50, and the type of resource is encoded by a one-hot vector.

\paragraph{Traffic Light Control}

We evaluate our approach in classic traffic light control problems.
In such problems the goal is usually to minimize the accumulated waiting time of all vehicles over all agents. 
We consider the accumulated waiting time for each agent at each intersection separately.
We use Simulation of Urban MObility (SUMO)\footnote{\url{https://github.com/eclipse/sumo}} to simulate a 3x3 intersection grid that has total of 9 agents where each agent controls the traffic light phase of one intersection.
Each intersection has 4 roads and a total of 8 lanes with different numbers of vehicles. 
A traffic light phase controls the traffic and specifies which lanes have the green light.
In our setting, we assume that each intersection have 4 phases. 
Depending on the type of intersections, different types of phase pattern can be considered.

The state space is composed of the current traffic light phase and for each lane, its queue and density of cars stopped at the intersection. 
An action corresponds to choose the current traffic light phase. 
The environment simulates intersections for 5000 seconds, with an episode length of 500 decision steps.
At each time step, new vehicles enter into the intersection with a fixed destination.

The minibatch size used is 50 and $\gamma = 0.99$.
The neighborhood definition is based on the position on the grid (the number of neighbors vary from 2 to 4).

The reward function is defined as the total waiting time at the current intersection subtracted from the accumulated waiting time at the last step. This definition of reward function motivates agent to minimize the accumulated waiting time which is our objective in this domain.
We have performed experiments in both cases where the reward is the accumulated waiting time for each agent ($D = N$) and where each agent's reward is further split into two components corresponds to two roads of the network ($D = 2N$). 

Note that in this simulation if the agents' decisions are too bad, a situation of complete blockage can occur. 
If this is the case, then regardless of the actions taken by the agents for the rest of the episode, traffic will remain blocked everywhere.

\begin{figure}[H]
    \centering
    \includegraphics[height=4.5cm]{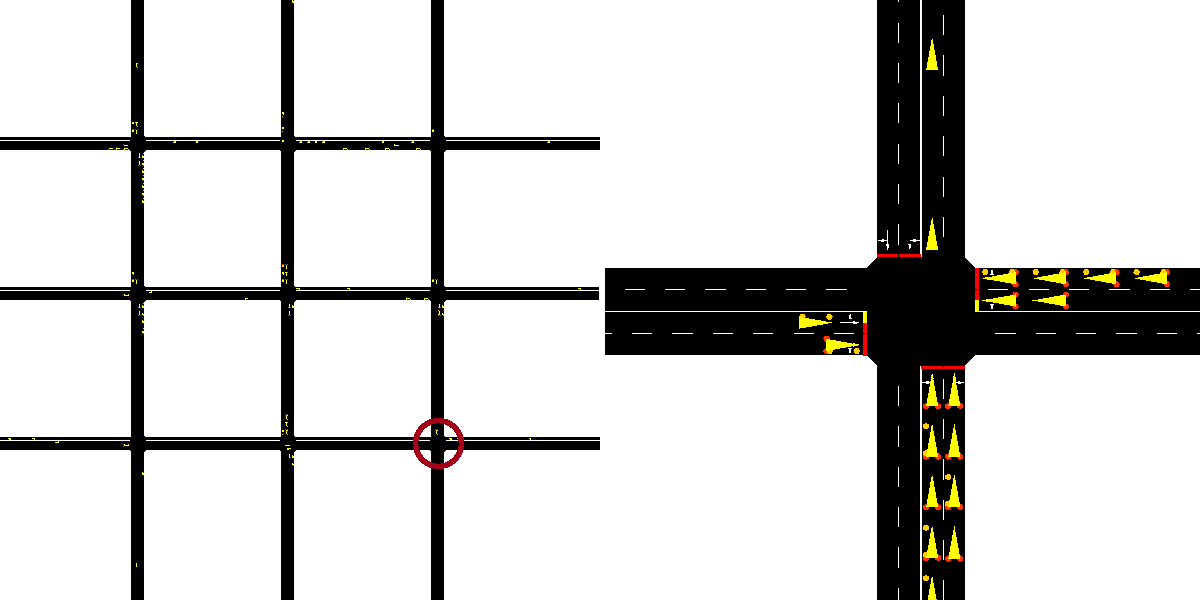}
    \caption{SUMO simulation with nine intersections in a grid. A zoom on one intersection is depicted at the bottom left. 
    Cars can arrive from any direction and go to any of the 3 remaining ones. }
    \label{fig:sumo_simuation}
\end{figure}

\begin{figure}[H]
    \centering
    \includegraphics[width=0.6\linewidth]{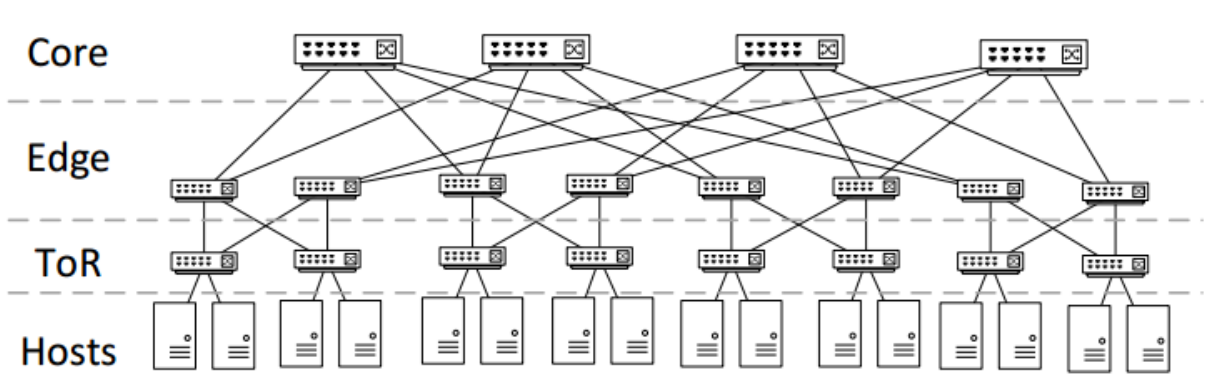}
    \caption{Network with a fat-tree topology from \citep{iroko}.}
    \label{fig:iroko_network}
\end{figure}

\paragraph{Data Center Control}
\us{In the Data Center Traffic Control problem, controllers manage the computer network that is shared by certain number of hosts. 
For the network topology, a fat-tree topology is considered (see Figure~\ref{fig:iroko_network}). 
In this topology, there are 16 hosts that are connected at the bottom with 20 switches.
Each switch is connected with 4 ports, which results in a total of 80 queues in the network.
We used Mininet\footnote{\url{https://github.com/mininet/mininet}} to simulate the network and Gobin\footnote{\url{https://github.com/udhos/goben}} to generate network traffic. 
In classic data center control problems, the goal is usually to maximize the sum of host bandwidths in order to avoid the network congestion.
However, we instead aim at maximizing the bandwidth of each host/agent in order to ensure fairness.

The minibatch size used is 128 and the $\gamma = 0.95$. 
\mz{We consider the neighbors as the 3 closest hosts which can be linked with a maximum of two switches.}
The total number of agents are 16, where each agent is dedicated to each host.

The global state is a $n \times m$ matrix, where $n$ is the number of ports in each switch and $m$ is the number of network features collected by Goben.
The continuous action corresponds to the allowed bandwidth for each host.

The $D$-dimensional reward vector is defined as follows:
    $$\vR_{\bm a,s} = \bm{a} - 2*\bm{a}*\max_{i}q_{i}(s)$$
where $\bm a$ is the vector action that represents the bandwidth allocation and  $q_i(s) $ represents the $i$-th queue length.
The reward is a vector, adapted from the \citep{iroko}, whose components are bandwidths per host penalized by the maximum of queue lengths.
In the original environment, the average of $\bm a$ is taken to define a scalar reward.

}

\end{document}